\documentclass[a4paper,fleqn]{cas-dc}

\usepackage[authoryear,longnamesfirst]{natbib}

\definecolor{bestcell}{RGB}{223,246,231}    
\definecolor{secondcell}{RGB}{255,244,214}   

\usepackage{xcolor}
\newcommand{\best}[1]{\cellcolor{bestcell}\textbf{#1}}
\newcommand{\second}[1]{\cellcolor{secondcell}#1}
\colorlet{upcol}{green!55!black}
\colorlet{downcol}{red!60!black}
\colorlet{flatcol}{gray!65!black}

\newcommand{\inc}[2]{#1\,{\raisebox{0.25ex}{\scriptsize\textcolor{upcol}{$\uparrow$\,#2}}}}
\newcommand{\dec}[2]{#1\,{\raisebox{0.25ex}{\scriptsize\textcolor{downcol}{$\downarrow$\,#2}}}}
\newcommand{\same}[1]{#1\,{\raisebox{0.25ex}{\scriptsize\textcolor{flatcol}{$\rightarrow$\,0}}}}

\def\tsc#1{\csdef{#1}{\textsc{\lowercase{#1}}\xspace}}
\tsc{WGM}
\tsc{QE}

\begin{document}
\let\WriteBookmarks\relax
\def\floatpagepagefraction{1}
\def\textpagefraction{.001}

\shorttitle{}    

\shortauthors{}  

\title [mode = title]{Hierarchical Dual-Change Collaborative Learning for UAV Scene Change Captioning}  



%


\author[1]{Fuhai Chen}
\author[1]{Pengpeng Huang}
\author[1]{Junwen Wu}
\author[1]{Hehong Zhang}
\author[1]{Shiping Wang}
\author[3]{Xiaoguang Ma}
\author[2]{Xuri Ge}
\cormark[1]
\ead{xurigexmu@gmail.com}

\affiliation[1]{organization={Fuzhou University},
            city={Fuzhou},
            country={China}}

\affiliation[2]{organization={Shandong University},
            city={Jinan},
            country={China}}

\affiliation[3]{organization={Foshan Graduate School of Innovation, Northeastern University},
            city={Foshan},
            country={China}}

\cortext[1]{Corresponding author}













\fntext[1]{}


\begin{abstract}
This paper proposes a novel task for UAV scene understanding - UAV Scene Change Captioning (UAV-SCC) - which aims to generate natural language descriptions of semantic changes in dynamic aerial imagery captured from a movable viewpoint.
Unlike traditional change captioning that mainly describes differences between image pairs captured from a fixed camera viewpoint over time, UAV scene change captioning focuses on image-pair differences resulting from both temporal and spatial scene variations dynamically captured by a moving camera.
The key challenge lies in understanding viewpoint-induced scene changes from UAV image pairs that share only partially overlapping scene content due to viewpoint shifts caused by camera rotation, while effectively exploiting the relative orientation between the two images.
To this end, we propose a Hierarchical Dual-Change Collaborative Learning (HDC-CL) method for UAV scene change captioning.
In particular, a novel transformer, \emph{i.e.} Dynamic Adaptive Layout Transformer (DALT) is designed to adaptively model diverse spatial layouts of the image pair, where the interrelated features derived from the overlapping and non-overlapping regions are learned within the flexible and unified encoding layer.
Furthermore, we propose a Hierarchical Cross-modal Orientation Consistency Calibration (HCM-OCC) method to enhance the model's sensitivity to viewpoint shift directions, enabling more accurate change captioning.
To facilitate in-depth research on this task, we construct a new benchmark dataset, named UAV-SCC dataset, for UAV scene change captioning. Extensive experiments demonstrate that the proposed method achieves state-of-the-art performance on this task. The dataset and code will be publicly released upon acceptance of this paper.
\end{abstract}


\begin{highlights}
\item We introduce a novel task, UAV Scene Change Captioning (UAV-SCC), which aims to describe semantic changes in dynamic aerial imagery captured from moving UAV viewpoints.
\item We propose a Hierarchical Dual-Change Collaborative Learning (HDC-CL) framework with a Dynamic Adaptive Layout Transformer (DALT) and a Hierarchical Cross-modal Orientation Consistency Calibration (HCM-OCC) strategy to model spatial layout variation and cross-modal semantic consistency.
\item We construct a benchmark UAV-SCC dataset and demonstrate the state-of-the-art performance of the proposed method through extensive experiments.
\end{highlights}



\maketitle

\section{Introduction}

The rapid evolution of Unmanned Aerial Vehicles (UAVs) has enabled a wide range of applications across diverse domains, ranging from environmental monitoring to autonomous infrastructure inspection.
These advanced applications, however, inherently demand efficient and lightweight data processing solutions for backend analysis, event logging, or human verification of environmental changes. When relying on the capture and transmission of sequential visual data (\emph{e.g.}, video streams), this process often encounters significant challenges due to high spatiotemporal redundancy, including low efficiency in manual retrospective verification, severe transmission latency, and excessive storage consumption~\cite{song2024knowledge,fang2025task,jiang2025improved}. These limitations critically restrict the broader and more intelligent application of UAV technology~\cite{ahmad2025future,igwenagu2025integrated,zhang2025uavdetr}.
In addition, recent studies have explored efficient deep learning techniques to reduce the computational overhead of visual models. Model compression approaches, such as network quantization, have been widely investigated to reduce the computational cost and memory footprint of deep neural networks while maintaining competitive performance~\cite{zhong2025mbquant}.
A promising alternative is to generate concise natural language descriptions that encapsulate essential scene semantics. Such textual representations enable humans to rapidly comprehend observed changes while serving as a lightweight substitute for conventional video or image data, substantially reducing communication and storage overhead without sacrificing critical information.

\begin{figure}
	\centering
	\includegraphics[width=\columnwidth]{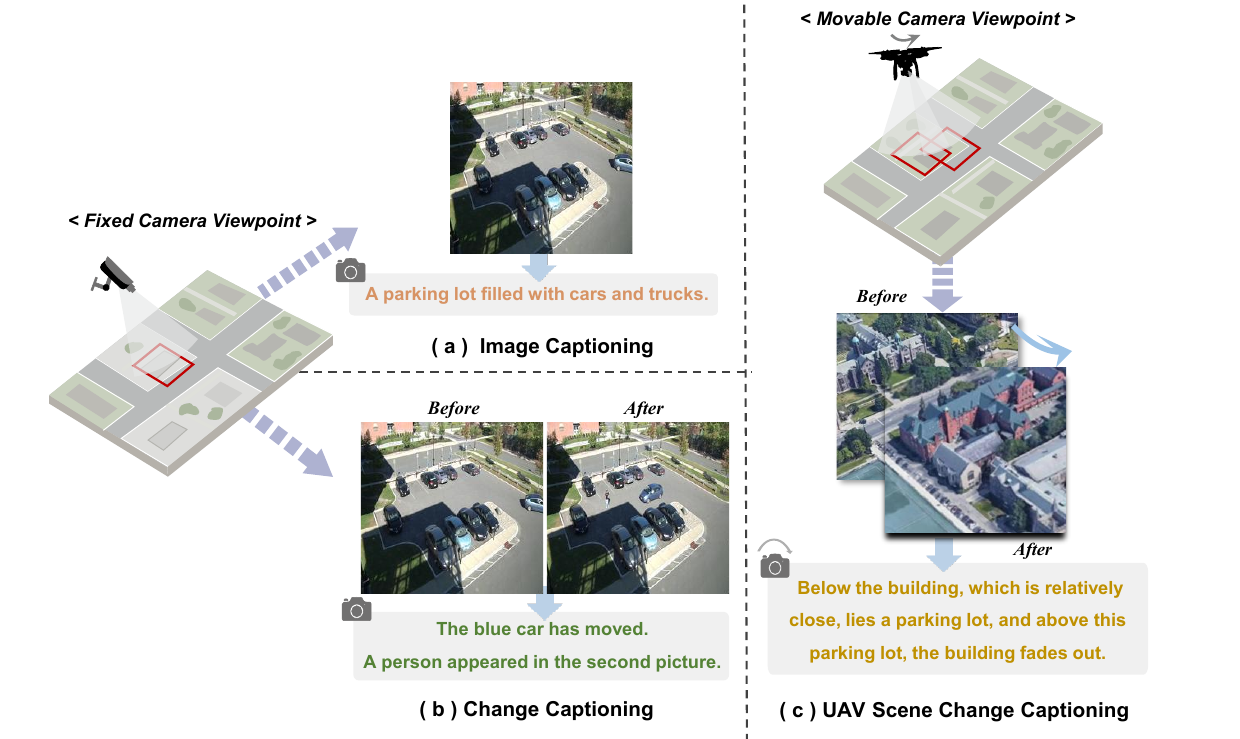}
	\caption{Comparison of three image captioning tasks. 
(a) Image Captioning generates semantic descriptions for a single image captured from a fixed camera viewpoint. 
(b) Change Captioning produces textual descriptions of differences between two images captured from the same viewpoint. 
(c) UAV Scene Change Captioning generates natural language descriptions of scene variations between image pairs captured from moving camera viewpoints. 
The gray boxes show example captions generated for each task.}
	\label{FIG:1}
\end{figure}

Inspired by image captioning (IC)~\cite{xu_altogether_2024,cai2024top,ramakrishnan_rona_2025,liu2025synthesize,jiang2025relational} that describes the semantic content of an image via natural language (Figure~\ref{FIG:1}(a)), this paper proposes a novel task, \emph{i.e.}, \textbf{UAV} \textbf{S}cene \textbf{C}hange \textbf{C}aptioning \textbf{(UAV-SCC)}, which aims to generate textual descriptions that encapsulate scene changes between image pairs captured from a movable UAV camera. 
Distinct from change captioning ~\cite{park_robust_2019,tu_semantic_2021,tu_selfsupervised_2023,tu_smart_2024,tu_distractorsimmune_2024,tu_contextaware_2024,yang2025restricted}, a task evolved from IC, which also conveys scene object-level semantic changes between image pairs (Figure~\ref{FIG:1}(b)), UAV-SCC must contend with scene variations due to the camera's viewpoint motion instead of static viewpoint under the fixed camera (Figure~\ref{FIG:1}(c)).
In such scenarios, the two images often share only partially overlapping scene content, resulting in inconsistent spatial layouts.
This setting introduces two major challenges. First, the model must effectively capture the relationships between overlapping and non-overlapping regions across the image pair, which requires robust alignment and comparison mechanisms to handle parallax effects. Second, it must capture directional cues introduced by viewpoint motion, which are critical for correctly interpreting and describing the observed scene variations.

However, existing change captioning methods remain limited when applied to UAV scenarios with dynamic viewpoints. Although recent works have introduced mechanisms such as cross-view alignment to mitigate viewpoint discrepancies, these approaches are typically developed for image pairs with largely aligned scenes. In contrast, UAV imagery captured from moving platforms often exhibits significant viewpoint shifts, resulting in partially overlapping scene content and inconsistent spatial layouts between paired images. Under such conditions, accurately distinguishing overlapping and non-overlapping regions becomes challenging. Moreover, the directional cues introduced by viewpoint motion play a crucial role in interpreting scene variations, yet they are rarely modeled explicitly in existing approaches. Consequently, these methods may produce ambiguous captions that fail to precisely describe the observed scene changes.

To address the aforementioned challenges, we propose a novel method, \emph{i.e.} \textbf{H}ierarchical \textbf{D}ual-\textbf{C}hange \textbf{C}ollaborative \textbf{L}earning (\textbf{HDC-CL}) for UAV scene change captioning. 
Specifically, to effectively capture the relationships between overlapping and non-overlapping regions, we first introduce the Dynamic Adaptive Layout Transformer (DALT), which enables adaptive modeling of diverse spatial layouts of image pairs. Considering the parallax effects caused by viewpoint shifts in UAV imagery, DALT incorporates a shift voting mechanism to estimate the overlapping regions between paired images. 
Based on this design, HDC-CL learns common and differential representations of image pairs and scene-level semantic changes through hierarchical consistency constraints and an independence regularization strategy. The scene change representation is further enhanced by distilling both local and global difference features from the paired images. Finally, we design a Hierarchical Cross-modal Orientation Consistency Calibration (HCM-OCC) strategy to improve the learning of viewpoint direction cues by aligning visual change features with directional textual semantics through a cross-modal consistency constraint.

\noindent The contributions are as follows:
\begin{itemize}
\item We propose a novel task, UAV Scene Change Captioning (UAV-SCC), which aims to generate natural language descriptions of scene variations in dynamic aerial imagery. In contrast to conventional change captioning tasks that typically operate on image pairs with largely aligned scenes, this task involves scene changes caused by camera motion.
\item We propose a Hierarchical Dual-Change Collaborative Learning (HDC-CL) framework that integrates a Dynamic Adaptive Layout Transformer (DALT) for modeling spatial layout variations and a Hierarchical Cross-modal Orientation Consistency Calibration (HCM-OCC) strategy for capturing directional semantics of viewpoint shifts.
\item We construct a benchmark dataset for UAV scene change captioning and conduct extensive experiments to validate the effectiveness of the proposed method, demonstrating state-of-the-art performance on this task.
\end{itemize}

\section{Related Work}
\label{sec:formatting} 

\subsection{Change captioning}

Image captioning has long been regarded as a fundamental task for visual understanding~\cite{xu_altogether_2024,ramakrishnan_rona_2025}, aiming to generate natural language descriptions that summarize the semantic content of an image.
Recent advances in large multimodal models have further improved performance on various vision–language tasks.
While large multimodal models offer strong generalization, their heavy computational requirements and the real-time visual transmission pressure make them impractical for both UAV-side and edge-cloud deployment. To address these limitations, automated architecture design techniques have also been explored to improve the efficiency of visual models, enabling better trade-offs between computational cost and model performance~\cite{zheng2022neural}.

Change captioning extends image captioning by generating natural language descriptions that explain visual differences between paired images. Change captioning extends image captioning by generating natural language descriptions that explain visual differences between paired images. The task was first introduced with the \textit{Spot-the-Diff} dataset~\cite{jhamtani_learning_2018a}, which contains real-world image pairs with sentence-level descriptions of scene differences. Later, the synthetic \textit{CLEVR-Change} dataset~\cite{park_robust_2019} provided controlled scene modifications and diverse change types, enabling systematic evaluation of change captioning models.
Early methods mainly focused on detecting single object-level changes by computing pixel-level or feature-level differences between paired images~\cite{park_robust_2019,liao2021sg}. However, these approaches rely heavily on spatial alignment and perform poorly under viewpoint variation. To improve robustness, subsequent work leveraged local feature matching~\cite{shi2020mvam, tu2021r3net} or contrastive learning~\cite{tu_selfsupervised_2023} to enhance change representation across views.
More recent efforts target multi-change captioning, where multiple changes are described within a single image pair. For example, IFDC~\cite{huang_image_2021} combines fine-grained visual, semantic, and positional cues at the instance level, while CARD~\cite{tu_contextaware_2024} explicitly separates shared context from difference-specific features to generate more accurate captions.
Despite these advances, most existing change captioning methods are developed under settings where paired images share largely aligned scene structures. In UAV imagery, however, viewpoint motion may lead to partially overlapping scene content and inconsistent spatial layouts, which introduces additional challenges for accurately modeling scene variations and generating reliable descriptions.

\subsection{UAV scene understanding tasks}

UAV scene understanding tasks aim to analyze dynamic environments using visual and multimodal data.
Early work primarily focused on static-scene tasks such as object detection~\cite{zhang2019Syolo,suo2023hit,li_remdet_2024,dong2024crop,li_selfprompting_2025,zhang2025uavdetr} and semantic segmentation~\cite{rizzoli2023syndrone}, relying on aerial imagery to localize and classify objects. However, these approaches often struggle with the scene dynamics inherent to UAV platforms.
With the rapid development of artificial intelligence techniques, UAV perception systems have been extended to more advanced applications, including visual tracking~\cite{zhang2024awesome}, scene monitoring~\cite{feng2024fc,igwenagu2025integrated}, and scene segmentation~\cite{ma2024unsupervised}.
In addition, multimodal fusion methods have been proposed to improve robustness in complex aerial environments by integrating information from multiple sensing modalities~\cite{fan2023avdn,su2023tggat,chu2024towards}.
Recent studies have also explored image generation and manipulation techniques to better model object interactions and scene consistency. For example, image editing frameworks such as Move-and-Act enable object-level manipulation while preserving background structures, providing useful insights for modeling complex scene variations in visual tasks~\cite{jiang2025move}.
Nevertheless, most existing UAV vision studies focus on perception-level tasks or global scene analysis, while language-based interpretation of dynamic scene changes remains relatively underexplored. In real-world UAV applications, where viewpoints change continuously and environments evolve over time, there is an increasing need for models that can semantically describe scene variations. This motivates the study of UAV scene change captioning.

\section{UAV Scene Change Captioning}

\subsection{Task Definition}
UAV Scene Change Captioning (UAV-SCC) is a multimodal task that aims to generate natural language descriptions of semantic differences between two images captured by a UAV at different times by moving its viewpoint over the scene.
Specifically, Given a pair of aerial images captured before and after a scene change, denoted as \( \mathbf{I}_{\text{bef}} \) and \( \mathbf{I}_{\text{aft}} \), the goal of UAV-SCC is to generate a captioning \( \mathbf{C} = \{ c_1, c_2, \ldots, c_m \} \) that accurately reflects the semantic changes between the two images.

Unlike conventional change captioning tasks where the image pairs are captured from a fixed viewpoint and typically exhibit pixel-level alignment, UAV-SCC introduces viewpoint variation and parallax error. The former results in partially overlapping of image pairs with partially common scene context, while the latter leads to pixel-level inconsistency of the partially overlapping area.


\begin{table}[t]
    \centering
    \fontsize{9}{11}\selectfont
    \renewcommand\tabcolsep{3pt}
    \caption{Statistics of the UAV-SCC dataset. UAV-SCC\textsubscript{Simple} and UAV-SCC\textsubscript{Rich} share similar data organization but differ in annotation volume and linguistic diversity.}
    \vspace{-0.5em}
    \label{dataset}
    \begin{tabular}{c|ccc|ccc}
        \toprule
        \multirow{2}{*}{\textbf{Data}} &
        \multicolumn{3}{c|}{\textbf{UAV-SCC\textsubscript{Simple}}} &
        \multicolumn{3}{c}{\textbf{UAV-SCC\textsubscript{Rich}}} \\
        \cmidrule(lr){2-4}\cmidrule(lr){5-7}
         & \textbf{Train} & \textbf{Val} & \textbf{Test} & \textbf{Train} & \textbf{Val} & \textbf{Test} \\
        \midrule
        Num. of image pairs & 7213 & 901 & 903 & 5643 & 705 & 706 \\
        Avg. caption length & 27.46 & 27.81 & 27.21 & 14.31 & 14.30 & 14.23 \\
        \bottomrule
    \end{tabular}
    \vspace{-0.8em}
\end{table}

\begin{figure*}[t]
  \centering
  \includegraphics[width=\textwidth]{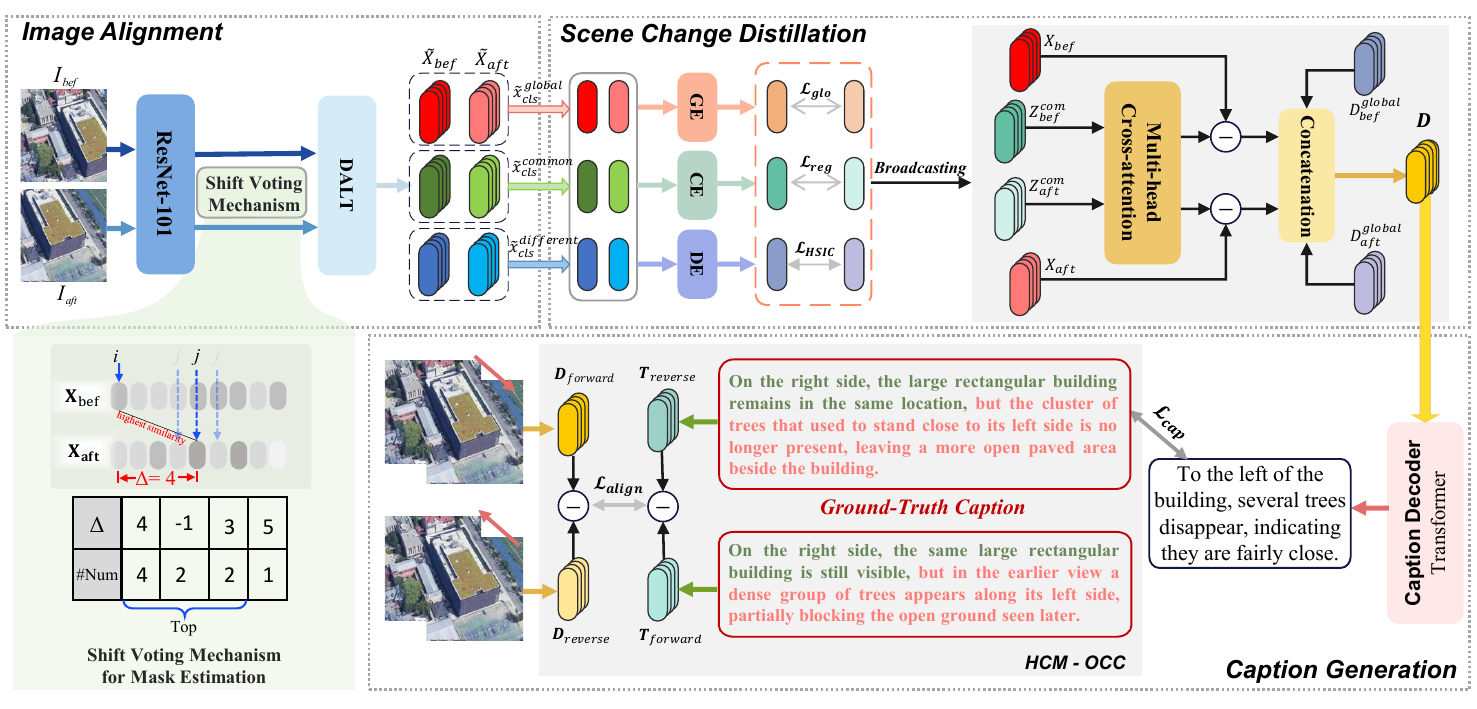}
  \caption{Overview of the HDC-CL framework. It consists of three steps: 
(1)~\textit{image alignment} for aligning the before–after image pair to correct spatial shifts and ensure consistent correspondence between regions,
(2)~\textit{scene change distillation}, where GE, CE, and DE denote the global, common, and different encoders,
and (3)~\textit{caption generation} based on distilled features.}
  \label{fig:figure2}
\end{figure*}

\begin{figure}
	\centering
	\includegraphics[width=\columnwidth]{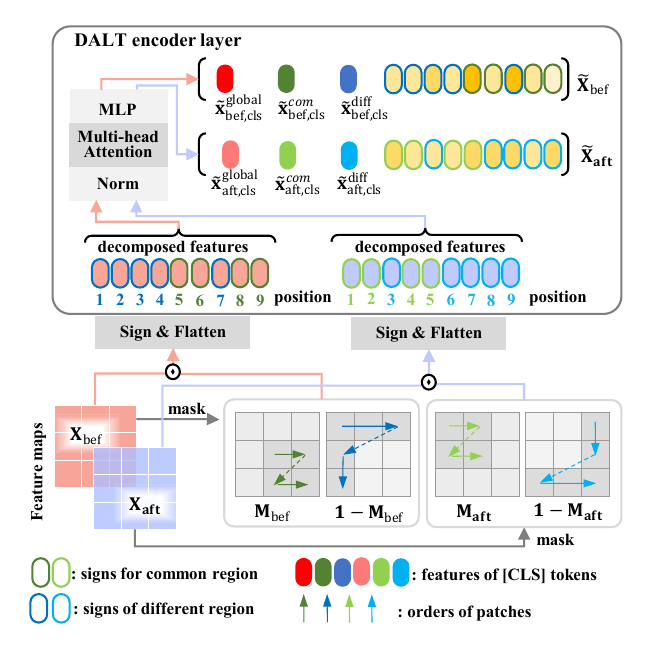}
	\caption{An illustration of the Dynamic Adaptive Layout Transformer (DALT) pipeline. For clarity, this example uses \( N = 9 \) patches per image.}
	\label{fig:figure3}
\end{figure}

\subsection{UAV Scene Change Captioning Dataset}
Existing change captioning datasets generally fall into two categories:
(1) real-world image pairs captured by static surveillance systems or UAVs operating under fixed viewpoints, typically consisting of consecutive frames from fixed-perspective video footage~\cite{jhamtani_learning_2018a,wu2024anticipation}, and
(2) synthetic image pairs generated using simulation environments~\cite{park_robust_2019}.
However, these datasets either assume fixed viewpoints or rely on simplified backgrounds, limiting their applicability to real-world UAV scenarios.

To address these limitations, we construct a new benchmark dataset with two versions, \emph{i.e.} the \textbf{UAV} \textbf{S}cene \textbf{C}hange \textbf{C}aptioning \textbf{(UAV-SCC)} datasets. 
Specifically, the image pairs are constructed by integrating UAV imagery from the publicly available datasets, GeoText-1652~\cite{chu2024towards} and UAVDT~\cite{du2018unmanned}.
We extract raw UAV images from both datasets and form scene-change pairs.
It is worth noting that only the original images from GeoText-1652 and UAVDT are used, while all scene-change pairs and corresponding captions are newly curated for our benchmark.
Based on the image pairs, we release two annotation versions, \emph{i.e.} UAV-SCC\textsubscript{Simple} and UAV-SCC\textsubscript{Rich} that differ in linguistic breadth and granularity. UAV-SCC\textsubscript{Simple} provides three captions per pair with concise, spatial-relation–oriented descriptions and relatively uniform lexical and syntactic patterns. In contrast, UAV-SCC\textsubscript{Rich} provides five captions per pair with substantially higher linguistic diversity, offering varied phrasings and structures to express scene changes while maintaining factual consistency. Both versions are annotated by three experts from the UAV perspective and include forward (before→after) and reverse (after→before) descriptions under a unified guideline. The final dataset statistics are summarized in Table \ref{dataset}. The additional preprocessing details and representative examples can be found in the Supplementary Materials.

\section{Methodology}

\subsection{Overview}
\label{sec:4.1}

The overall architecture of our proposed method is illustrated in Figure~\ref{fig:figure2}. To address the unique challenges of scene change captioning in UAV scenarios, we propose a Hierarchical Dual-Change Collaborative Learning (HDC-CL) framework, which can be divided into three steps: (1) image alignment, (2) scene change distillation, and (3) caption generation. Specifically, we design a shift voting mechanism to automatically find out the common mask of overlapped parts on feature level and introduce a Dynamic Adaptive Layout Transformer (DALT) in the image alignment stage to effectively model both common and different regions between image pairs. In the caption generation stage, we further propose a Hierarchical Cross-Modal Orientation Consistency Calibration (HCM-OCC) module to enhance the model's sensitivity to viewpoint variations. 

\subsection{Image Alignment}
\label{4.2}
Specifically, given a pair of UAV images \( \mathbf{I}_{\text{bef}} \) and \( \mathbf{I}_{\text{aft}} \), we first leverage an off-the-shelf image encoder ResNet-101~\cite{he2016resnet} to extract local visual features of \( N \) patches from each image, resulting in the initial feature sequence \( \mathbf{X}_o = \{ x_o^1, x_o^2, \ldots, x_o^N \} \), where \( x_o^i \in \mathbb{R}^d \) and \( o \in \{\text{bef}, \text{aft}\} \)  denotes the "before" and "after" images. We also define the set of region types \( \mathcal{R} = \{\text{glo},\ \text{com},\ \text{diff}\} \) denotes overall image, overlapping area, nonoverlapping areas, respectively.

\subsubsection{Shift Voting Mechanism for Mask Estimation}

Due to the parallax error between the images from moving camera, UAV-SCC needs to adaptively align the image feature for robustness.
To this end, we design \textit{shift voting mechanism} to adaptively estimate the mask of the overlapping area based on the patch-level feature map.

Specially, given the flattened token lists of two images \( \mathbf{I}_{\text{bef}} \) and \( \mathbf{I}_{\text{aft}} \) as Figure~\ref{fig:figure2}, for the $i$-th token of \( \mathbf{I}_{\text{bef}} \), this mechanism first computes $(i,j)$ pairwise feature similarities $s_{ij}$ for each $j$-th token in \( \mathbf{I}_{\text{aft}} \). Subsequently, it finds out the maximal $s_{ij}$ for the $i$-th token of \( \mathbf{I}_{\text{bef}} \) and calculates the $j$-to-$i$ relative location $\Delta$. Then, this mechanism counts the number of different $\Delta$ values across different $i$ and votes the top-ranked $\Delta$ values as candidate offsets of the common regions across different images.

We finally filter the $\Delta$ values by $S(\Delta)$, the sum of the similarities of all token pairs under such each $\Delta$ value. For each $\Delta$ value, we have
\begin{equation}
    S(\Delta) = \sum_{i,j} \max(s_{ij},0)\,\mathbf{1}(\Delta_{ij} = \Delta),
    \label{eq:vote_map}
\end{equation}
where $\mathbf{1}(\cdot)$ is the indicator function. 
The dominant relative location $\Delta^\star$ is then obtained by 
\begin{equation}
    \Delta^\star = \arg\max_{\Delta} S(\Delta).
    \label{eq:dominant_shift}
\end{equation}

In this way, the 2D mask of common region can be estimated based on 1D shifting value $\Delta^\star$ of the common tokens.

\subsubsection{Dynamic Adaptive Layout Transformer}
To model region-level semantic correspondence between the image pair, we propose the Dynamic Adaptive Layout Transformer (DALT), as illustrated in Figure~\ref{fig:figure3}. 
This module explicitly decomposes each image into \textit{common} and \textit{different} regions based on the masks estimated above, enhancing the model's ability to capture fine-grained structural changes.

Specially, we start from the estimated binary common mask 
\(\mathbf{M}_o \in \{0,1\}^N\), 
where \(\mathbf{M}_o^i = 1\) indicates that the \(i\)-th patch belongs to a semantically common region between the two views. 
We project this mask onto the feature map space and decompose the original features \(\mathbf{X}_o\) as follows:
\begin{equation}
\mathbf{X}_o^{\text{com}} = \mathbf{M}_o \cdot \mathbf{X}_o, \quad \mathbf{X}_o^{\text{diff}} = (1 - \mathbf{M}_o) \cdot \mathbf{X}_o,
\label{eq:mask_decomposition}
\end{equation}

This sequence is decomposed into region-specific subsets, and for each region type \( r \in \mathcal{R} \) is assigned a learnable [CLS] token, denoted as \( \mathbf{x}_{\text{cls}}^r \in \mathbb{R}^d \).
We define a sign set \( \mathcal{I}_o^r \subset \{1, 2, \dots, N\} \), which records the ordered patch indices in image \( \mathbf{X}_o \) that belong to region type \( r \). 
These sequences are subsequently fed into a multi-head self-attention~\cite{vaswani2017attention} encoder to extract context-enhanced semantic representations at both the global and regional levels.
As a result, we obtain updated region-aware features \( \widetilde{\mathbf{X}}_o = \{ \widetilde{\mathbf{x}}_{\text{o,cls}}^r,\ \widetilde{\mathbf{x}}_o^i \mid r \in \mathcal{R},\ i \in \mathcal{I}_o^r \} \), 
where \( \widetilde{\mathbf{x}}_{\text{o,cls}}^r \) denotes the aggregated representation for region type \( r \), and 
\( \widetilde{\mathbf{x}}_o^i \) is the context-enhanced feature for patch \( i \). Further implementation details and derivations are provided in the Supplementary Materials.


\subsection{Scene Change Distillation}
\label{4.3}
\paragraph{Context feature decoupling}
To capture multi-level semantics, we introduce region-specific context encoding for three types of regions. 
Specifically, we design two MLP encoders: a global encoder \( \mathrm{GE}(\cdot\,; \theta_G) \) for global regions and a common encoder \( \mathrm{CE}(\cdot\,; \theta_C) \) for common regions. 
For the different regions, we apply two different encoders \( \mathrm{DE}_{\text{bef}}(\cdot\,; \theta_{\text{bef}}) \) and \( \mathrm{DE}_{\text{aft}}(\cdot\,; \theta_{\text{aft}}) \). 
All encoders are implemented as lightweight linear projections.
We feed the [CLS] token of each region into its corresponding encoder to obtain the region-level context features:
\begin{equation}
\begin{aligned}
g_o &= \mathrm{GE}(\widetilde{\mathbf{x}}_{o,\text{cls}}^{\text{glo}};\theta_G), \\
c_o &= \mathrm{CE}(\widetilde{\mathbf{x}}_{o,\text{cls}}^{\text{com}};\theta_C), \\
d_o^{global} &= \mathrm{DE}_o(\widetilde{\mathbf{x}}_{o,\text{cls}}^{\text{diff}};\theta_o),
\end{aligned}
\end{equation}
where \( g_o \), \( c_o \), and \( d_o^{global} \in \mathbb{R}^d \) denote the global, common, and different context representations.

\paragraph{Hierarchical consistency constraints}
To better capture stable semantics in image pairs, we design two levels of contrastive objectives: a \textit{global consistency constraint} and a \textit{common region consistency constraint}.

The global consistency constraint aims to align the overall background semantics of the image pair, which often remain unchanged in UAV imagery. Meanwhile, the common region consistency constraint focuses on the overlapping regions with unchanged objects, encouraging the model to learn consistent representations of semantically aligned regions.

Formally, we apply InfoNCE loss~\cite{oord2018representation} to the normalized global features \( \hat{g}_{\text{bef}}, \hat{g}_{\text{aft}} \) and common region features \( \hat{c}_{\text{bef}}, \hat{c}_{\text{aft}} \) within a batch of size \( B \):
\begin{equation}
\mathcal{L}_{\text{glo}} = - \frac{1}{B} \sum_{i=1}^{B} \log \frac{
\exp\left( \hat{g}^{(i)}_{\text{bef}} \cdot \hat{g}^{(i)}_{\text{aft}} / \tau \right)
}{
\sum_{j=1}^{B} \exp\left( \hat{g}^{(i)}_{\text{bef}} \cdot \hat{g}^{(j)}_{\text{aft}} / \tau \right)
},
\label{eq:global-nce}
\end{equation}

\begin{equation}
\mathcal{L}_{\text{reg}} = - \frac{1}{B} \sum_{i=1}^{B} \log \frac{
\exp\left( \hat{c}^{(i)}_{\text{bef}} \cdot \hat{c}^{(i)}_{\text{aft}} / \tau \right)
}{
\sum_{j=1}^{B} \exp\left( \hat{c}^{(i)}_{\text{bef}} \cdot \hat{c}^{(j)}_{\text{aft}} / \tau \right)
},
\label{eq:region-nce}
\end{equation}
where $\tau$ is the temperature factor. These two objectives capture complementary semantic information at different levels, enabling the model to learn aligned representations.

\paragraph{Independence regularization}  
To further disentangle the semantic representations of common and different regions, we introduce an independence regularization to reduce the statistical dependence between the difference context features from the before and after images. Specifically, we employ the Hilbert-Schmidt Independence Criterion~\cite{song2007supervised}(HSIC) to quantify the dependence between the normalized difference features \( \hat{d}_{\text{bef}}^{global} \) and \( \hat{d}_{\text{aft}}^{global} \).

We compute Gaussian kernel matrices \( \mathbf{K} \) and \( \mathbf{L} \) for \( \hat{d}_{\text{bef}}^{global} \) and \( \hat{d}_{\text{aft}}^{global} \), respectively. The HSIC loss is then defined as:
\begin{equation}
\mathcal{L}_{\text{HSIC}} = \frac{1}{(B - 1)^2} \operatorname{Tr}\left( \mathbf{K} \mathbf{H} \mathbf{L} \mathbf{H} \right),
\end{equation}
where \( \mathbf{H} = \mathbf{I} - \frac{1}{B} \mathbf{1}\mathbf{1}^\top \) is the centering matrix. \textbf{I} is an identity matrix and \( \mathbf{1} \) is an all-one column vector. This loss encourages the difference features to capture diverse and independent change information across the image pair. The final context distillation loss combines all three objectives:
\begin{align}
\mathcal{L}_{\text{con}} = \mathcal{L}_{\text{glo}} + \mathcal{L}_{\text{reg}} + \mathcal{L}_{\text{HSIC}}. \label{eq:context-loss}
\end{align}

\paragraph{Scene change distilling}
To effectively extract true semantic changes, we employ a context-guided mechanism based on region-wise feature differences. 
To model the cross-image context within the common regions, we concatenate the local features \(\mathbf{X}_o^{\text{com}}\) with the corresponding [CLS] token \(c_o\) and project them to obtain the enhanced feature representation, denoted as \( \mathbf{Z}_{\text{o}}^{com} \in \mathbb{R}^{N \times d} \), where the local features of different regions are zero-masked.
We apply a multi-head cross-attention (MHCA) module to model inter-image correspondence:
\begin{equation}
\begin{aligned}
\widetilde{\mathbf{X}}_{\text{bef}}^{\text{com}} &= \text{MHCA}(\mathbf{Z}_{\text{bef}}^{\text{com}}, \mathbf{Z}_{\text{aft}}^{\text{com}}, \mathbf{Z}_{\text{aft}}^{\text{com}}), \\
\widetilde{\mathbf{X}}_{\text{aft}}^{\text{com}} &= \text{MHCA}(\mathbf{Z}_{\text{aft}}^{\text{com}}, \mathbf{Z}_{\text{bef}}^{\text{com}}, \mathbf{Z}_{\text{bef}}^{\text{com}}).
\end{aligned}
\end{equation}

Next, we compute residual features by subtracting the context-aligned overlapping region from the original features \( \mathbf{X}_o \in \mathbb{R}^{N \times d} \):

\begin{equation}
\begin{aligned}
\mathbf{D}_o^{\text{local}} = \phi(\mathbf{X}_o - \widetilde{\mathbf{X}}_{o}^{\text{com}})
\end{aligned}
\end{equation}

where \( \phi(\cdot) \) denotes a nonlinear projection.
Finally, we obtain the fused difference representation by applying a nonlinear transformation to the concatenation of the global and local difference features, \( \mathbf{D}_o = \left[ \mathbf{D}_o^{\text{global}} ; \mathbf{D}_o^{\text{local}} \right] \), where \( \mathbf{D}_o^{\text{global}} \) is got by repeating \( d_o\), and \( [\cdot\ ;\ \cdot] \) represents feature concatenation.
Next, pass them through a nonlinear transformation to obtain the fused change representation:
\begin{equation}
\mathbf{D} = \text{ReLU}\left( \mathbf{D}_o \mathbf{W}_c + \mathbf{b}_c \right) \in \mathbb{R}^{N \times d}.
\end{equation}

This final representation \( \mathbf{D} \) integrates contextual cues and local differences, serving as a unified embedding of visual changes for downstream caption generation.

\subsection{Caption Generation}
\label{4.4}

After obtaining the unified difference representation \( \mathbf{D} \in \mathbb{R}^{N \times d} \), we employ a Transformer decoder~\cite{vaswani2017attention} to generate directional scene change captions. Specifically, given the token sequence of the ground-truth caption, we first compute its word embeddings and add positional encoding. These embeddings are passed through masked self-attention layers to capture intra-sentence dependencies, followed by cross-attention with \(\mathbf{D}\) to align each word with relevant visual changes. The enhanced contextual representations are then projected to the vocabulary space for word prediction:
\begin{equation}
\mathbf{S} = \text{Softmax}\left( \widehat{\mathbf{H}} \mathbf{W}_s + \mathbf{b}_s \right),
\end{equation}
where \(\widehat{\mathbf{H}} \in \mathbb{R}^{m \times d}\) is the decoded hidden representation of length \(m\), and \(\mathbf{W}_s \in \mathbb{R}^{d \times u}\), \(\mathbf{b}_s \in \mathbb{R}^{u}\) are learnable parameters with vocabulary size \(u\).

\textbf{Hierarchical cross-modal orientation consistency calibration.}
To enhance vision-language alignment, we introduce a cross-modal orientation consistency mechanism that explicitly models semantic directionality across modalities.

We first compute the visual directional vector by subtracting the reverse image feature from the forward one, \emph{i.e.}, 
\( \Delta \mathbf{d} = Aggregation(\mathbf{D}_{\text{forward}} - \mathbf{D}_{\text{reverse}}) \in \mathbb{R}^{d} \), where \( \mathbf{D}_{\text{forward}}, \mathbf{D}_{\text{reverse}} \in \mathbb{R}^{N \times d} \) are computed as \( \mathbf{D}\) respectively with forward and reverse orders.
Similarly, for the language modality, we encode the forward and reverse textual descriptions via the Transformer and obtain the textual directional vector as 
\( \Delta \mathbf{t} = Aggregation(\mathbf{T}_{\text{forward}} - \mathbf{T}_{\text{reverse}}) \in \mathbb{R}^{d} \).
We then apply a bidirectional margin ranking loss to encourage alignment between visual and textual semantics:

\begin{equation}
\begin{aligned}
\mathcal{L}_{\text{align}} = \frac{1}{2} \big[ 
& \max(0, \gamma - \cos(\Delta \mathbf{t}, \Delta \mathbf{d}) + \cos(\Delta \mathbf{t}, \Delta \mathbf{d}^{-})) \\
& \hspace*{-2em} + \max(0, \gamma - \cos(\Delta \mathbf{d}, \Delta \mathbf{t}) + \cos(\Delta \mathbf{d}, \Delta \mathbf{t}^{-})) 
\big],
\end{aligned}
\end{equation}
where the superscript “\(-\)” denotes negative sample randomly selected beyond this image pair, and \(\gamma\) is a margin hyperparameter.

\subsection{Model Training}
\label{sec:4.4}
Our method adopts an end-to-end training strategy by maximizing the consistency between the generated caption and the ground-truth caption to optimize model performance. Specifically, given a ground-truth sentence sequence \(\{s_1^*, s_2^*, \ldots, s_m^*\}\), we minimize the negative log-likelihood (NLL) loss as follows:
\begin{equation}
\mathcal{L}_{\text{cap}}(\theta) = -\sum_{t=1}^{m} \log p_\theta(s_t^* | s_{<t}^*),
\end{equation}
where \(p_\theta(s_t^* | s_{<t}^*)\) represents the probability of generating the \(t\)-th word given the previous ground-truth tokens, and \(\theta\) denotes all learnable parameters in the model. The overall objective function of our model is defined as a weighted sum of the caption generation loss, the context contrastive loss, and the bidirectional alignment loss:
\begin{equation}
\mathcal{L} = \mathcal{L}_{\text{cap}} + \lambda (\mathcal{L}_{\text{con}} + \mathcal{L}_{\text{align}}).
\end{equation}
where \(\lambda\) are hyperparameters that balance the importance of each loss component. The detailed settings of these weights are provided in the Supplementary Materials.

\begin{table*}[t]
    \centering
    \renewcommand{\arraystretch}{0.8}       
    \small   
    \caption{
Comparison with existing methods on the UAV-SCC\textsubscript{Simple} and UAV-SCC\textsubscript{Rich} datasets.
B: BLEU-4, M: METEOR, R: ROUGE-L, C: CIDEr, S: SPICE.
\colorbox{bestcell}{\textbf{GREEN}} indicates the best of all methods, and 
\colorbox{secondcell}{\textbf{YELLOW}} indicates the second best.
}
    \resizebox{\textwidth}{!}{
    \begin{tabular}{l|ccccc|ccccc}
        \toprule
        \multirow{2}{*}{\textbf{Method}} &
        \multicolumn{5}{c|}{\textbf{UAV-SCC\textsubscript{Simple}}} &
        \multicolumn{5}{c}{\textbf{UAV-SCC\textsubscript{Rich}}} \\
        \cmidrule(lr){2-6}\cmidrule(lr){7-11}
         & \textbf{B} & \textbf{M} & \textbf{R} & \textbf{C} & \textbf{S} 
         & \textbf{B} & \textbf{M} & \textbf{R} & \textbf{C} & \textbf{S} \\
        \midrule
        DUDA~\cite{park_robust_2019} & 7.36 & 17.69 & 24.39 & -- & 21.70 & 4.17 & 8.89 & 18.32 & -- & \best{14.19} \\
        SRDRLL~\cite{tu_semantic_2021} & 8.28 & 18.78 & 24.27 & -- & 25.48 & 3.64 & 5.29 & 15.19 & --& 10.26 \\
        SCORER+CBR~\cite{tu_selfsupervised_2023} & \second{28.16} & 25.74 & 42.88 & 33.48 & 28.90 & \second{18.93} & 17.30 & 42.49 & 8.57 & 9.80 \\
        SMART~\cite{tu_smart_2024} & 25.08 & 24.55 & 40.78 & 30.13 &26.44 & 16.37 & \second{17.66} & 40.35 & 8.93 & 10.16 \\
        DIRL+CCR~\cite{tu_distractorsimmune_2024} & 24.99 & 23.81 & 40.20 & 30.43 & 26.61 & 16.88 & 15.98 & 42.13 & 8.77 & 11.41 \\
        CARD~\cite{tu_contextaware_2024} & 27.49 & \second{26.23} & \second{42.98} & \second{48.66} & \second{30.76} & 18.66 & 16.46 & \best{45.03} & \second{15.75} & 11.87 \\
        \midrule
        \textbf{HDC-CL (Ours)} & \best{31.13} & \best{27.34} & \best{44.58} & \best{54.68} & \best{33.09} 
        & \best{19.26} & \best{18.45} & \second{44.32} & \best{19.16} & \second{13.00} \\
        \bottomrule
    \end{tabular}
    }
    \label{table1}
\end{table*}

\begin{table*}[t]
    \centering
    \fontsize{9}{11}\selectfont
    \renewcommand\tabcolsep{6pt}
    \caption{
Joint ablation of loss components in HDC-CL on the UAV-SCC\textsubscript{Rich} and UAV-SCC\textsubscript{Simple} datasets.
\textbf{Bold} numbers indicate the best performance in each column.
}

    \begin{tabular}{c c c | c c c c c | c c c c c}
        \toprule
        \multirow{2}{*}{$\mathcal{L}_{\text{global}}$} &
        \multirow{2}{*}{$\mathcal{L}_{\text{region}}$} &
        \multirow{2}{*}{$\mathcal{L}_{\text{HSIC}}$} &
        \multicolumn{5}{c|}{\textbf{UAV-SCC\textsubscript{Rich}}} &
        \multicolumn{5}{c}{\textbf{UAV-SCC\textsubscript{Simple}}}
        \\
        \cmidrule(lr){4-8} \cmidrule(lr){9-13}
        & & &
        \textbf{B} & \textbf{M} & \textbf{R} & \textbf{C} & \textbf{S} &
        \textbf{B} & \textbf{M} & \textbf{R} & \textbf{C} & \textbf{S}
        \\
        \midrule
        $\times$ & $\times$ & $\times$ &
        19.20 & 16.30 & 45.01 & 13.56 & 12.49 &
        29.39 & 26.66 & 44.44 & 50.70 & 31.54
        \\
        $\checkmark$ & $\times$ & $\times$ &
        18.72 & 17.17 & 43.81 & 18.61 & 12.28 &
        28.65 & 26.80 & 42.89 & 48.52 & 30.28
        \\
        $\times$ & $\checkmark$ & $\times$ &
        18.90 & 16.64 & 45.20 & 16.51 & 12.41 &
        29.50 & 26.82 & 44.45 & 52.00 & 31.60
        \\
        $\times$ & $\times$ & $\checkmark$ &
        19.26 & 17.21 & \textbf{45.60} & 17.94 & 12.62 &
        28.55 & 26.84 & 43.48 & 50.73 & 31.72
        \\
        $\checkmark$ & $\checkmark$ & $\times$ &
        18.63 & 17.44 & 43.87 & 18.92 & 12.58 &
        29.75 & 27.14 & 43.88 & 50.29 & 32.34
        \\
        $\checkmark$ & $\times$ & $\checkmark$ &
        \textbf{19.81} & 16.13 & 45.02 & 13.78 & 12.47 &
        28.69 & 26.32 & 43.42 & 48.76 & 31.55
        \\
        $\times$ & $\checkmark$ & $\checkmark$ &
        18.81 & 17.57 & 44.82 & 18.73 & 12.64 &
        29.20 & 26.73 & 43.78 & 51.51 & 31.54
        \\
        $\checkmark$ & $\checkmark$ & $\checkmark$ &
        19.26 & \textbf{18.45} & 44.32 & \textbf{19.16} & \textbf{13.00} &
        \textbf{31.13} & \textbf{27.34} & \textbf{44.58} & \textbf{54.68} & \textbf{33.09}
        \\
        \bottomrule
    \end{tabular}
    \label{table33}
\end{table*}

\section{Experiments}
\subsection{Experimental setup}
\textbf{Evaluation metrics.} As in the change captioning literature~\cite{tu_contextaware_2024}, we follow standard protocols to run the evaluation on the proposed UAV-SCC dataset benchmark, including BLEU-4 (B)~\cite{papi2002bleu}, METEOR (M)~\cite{lavie2007meteor}, ROUGE-L (R)~\cite{lin2004rouge}, CIDEr (C)~\cite{vedantam2015cider}, and SPICE (S)~\cite{anderson2016spice}. These metrics assess both the fluency of the generated sentences and their semantic consistency with ground-truth descriptions. All results are computed using the official Microsoft COCO evaluation server~\cite{chen2015cococaptions}.

\textbf{Implementation details.} We utilize a pre-trained ResNet-101~\cite{he2016resnet} to extract local visual features, yielding a feature map of dimensions $1024 \times 16 \times 16$. These features are subsequently projected into a 512-dimensional embedding space. 
The feature dimension of the learnable [CLS] token is set to 512. The hidden size of the entire model is set to 512 as well. For the UAV-SCC\textsubscript{Simple} dataset, the vocabulary size and maximum sequence length are 1359 and 129, respectively. For the UAV-SCC\textsubscript{Rich} dataset, these values are 3456 and 90, respectively.
The model is trained with a batch size of 32 for a total of $1 \times 10^4$ iterations. Optimization is performed using the Adam optimizer~\cite{king2017adam} with an initial learning rate of $2 \times 10^{-4}$. More details can be found in the Supplementary Material. 

\subsection{Comparison with Existing Methods}
Since UAV-SCC task is new, we compare our method with the most related task—change captioning. Specifically, we select representative baselines including: DUDA~\cite{park_robust_2019}, SRDRL~\cite{tu_semantic_2021}, and SCORER+CBR~\cite{tu_selfsupervised_2023}, and state-of-the-art approaches such as SMART~\cite{tu_smart_2024}, DIRL+CCR~\cite{tu_distractorsimmune_2024}, and CARD~\cite{tu_contextaware_2024}. For a fair comparison, we reproduce these methods on the UAV-SCC dataset based on their published papers and released official codes, using the same number of samples as our method.


\begin{table*}[t]
  \centering
  \renewcommand{\arraystretch}{1.2}
  \caption{
Ablation study of the proposed Hierarchical Cross-modal Orientation Consistency Calibration (HCM-OCC) module 
on the UAV-SCC datasets. 
The symbols $\uparrow$ and $\downarrow$ indicate performance improvements or drops compared with the corresponding baselines.
}
  \resizebox{\textwidth}{!}{%
  \begin{tabular}{l|ccccc|ccccc}
    \toprule
    \multirow{2}{*}{\textbf{Method}} &
    \multicolumn{5}{c|}{\textbf{UAV-SCC\textsubscript{Simple}}} &
    \multicolumn{5}{c}{\textbf{UAV-SCC\textsubscript{Rich}}}\\
    \cmidrule(lr){2-6}\cmidrule(lr){7-11}
    & \textbf{B} & \textbf{M} & \textbf{R} & \textbf{C} & \textbf{S}
    & \textbf{B} & \textbf{M} & \textbf{R} & \textbf{C} & \textbf{S} \\
    \midrule
    w/o HCM-OCC  & 29.62 & 26.84 & 43.33 & 51.47 & 31.82 & 18.86 & 16.99 & 44.03 & 18.66 & 12.42 \\
    HDC-CL & \inc{31.13}{+1.51} & \inc{27.34}{+0.50} & \inc{44.58}{+1.25} & \inc{54.68}{+3.21} & \inc{33.09}{+1.27}
                 & \inc{19.26}{+0.40} & \inc{18.45}{+1.46} & \inc{44.32}{+0.29} & \inc{19.16}{+0.50} & \inc{13.00}{+0.58} \\
    \midrule
    CARD         & 27.49 & 26.23 & 42.98 & 48.66 & 30.76 & 18.66 & 16.46 & 45.03 & 15.75 & 11.87 \\
    CARD+HCM-OCC & \inc{28.81}{+1.32} & \inc{26.89}{+0.66} & \inc{43.74}{+0.76} & \dec{47.26}{-1.40} & \inc{32.11}{+1.35}
                 & \inc{21.48}{+2.82} & \inc{17.42}{+0.96} & \inc{47.15}{+2.12} & \inc{18.55}{+2.80} & \inc{12.62}{+0.75} \\
    DIRL         & 24.99 & 24.81 & 40.20 & 30.43 & 26.61 & 16.88 & 15.98 & 42.13 & 8.77  & 11.41 \\
    DIRL+HCM-OCC & \inc{25.67}{+0.68} & \same{24.81} & \inc{40.89}{+0.69} & \inc{31.09}{+0.66} & \inc{26.69}{+0.08}
                 & \dec{16.68}{-0.20} & \inc{16.90}{+0.92} & \dec{42.03}{-0.10} & \inc{8.99}{+0.22}  & \inc{11.99}{+0.58} \\
    \bottomrule
  \end{tabular}
  }
  \label{table4}
\end{table*}


Table~\ref{table1} presents the performance comparison between our proposed method and existing change captioning approaches on the UAV-SCC dataset. HDC-CL achieves the best overall performance on both UAV-SCC\textsubscript{Simple} and UAV-SCC\textsubscript{Rich} datasets.
Specifically, on the headline metric CIDEr, HDC-CL reaches 54.68 on UAV-SCC\textsubscript{Simple} and 19.16 on UAV-SCC\textsubscript{Rich}, outperforming the second-best method CARD by 6.02 and 3.41, respectively.
These substantial gains on the task’s primary evaluation metric clearly demonstrate that HDC-CL can generate more semantically contextually precise change descriptions under both concise and linguistically diverse conditions.








\begin{table*}[t]
\centering
\footnotesize
\setlength{\tabcolsep}{3.5pt}
\renewcommand{\arraystretch}{1.05}
\caption{
Ablation study of DALT under different mask generation strategies on the UAV-SCC datasets.
\textbf{Bold} numbers indicate the best performance in each column.
}
\begin{tabular}{l|ccccc|ccccc}
\toprule
\multirow{2}{*}{\textbf{Mask strategy variants}} &
\multicolumn{5}{c}{\textbf{UAV-SCC\textsubscript{Simple}}} &
\multicolumn{5}{c}{\textbf{UAV-SCC\textsubscript{Rich}}} \\
\cmidrule(lr){2-6}\cmidrule(lr){7-11}
& \textbf{B} & \textbf{M} & \textbf{R} & \textbf{C} & \textbf{S}
& \textbf{B} & \textbf{M} & \textbf{R} & \textbf{C} & \textbf{S} \\
\midrule
w/o Mask
& 25.60 & 25.56 & 41.27 & 46.10 & 29.97
& 18.22 & 17.09 & 43.54 & 18.66 & 12.31 \\

Random Mask
& 28.00 & 26.48 & 43.31 & 52.12 & 31.26
& 18.41 & 17.21 & 42.95 & 18.33 & 12.08 \\

Random Direction Window
& 29.68 & 26.89 & 43.80 & 52.20 & 31.95
& 18.70 & 17.41 & 42.85 & 18.51 & 12.74 \\

No Direction Window
& 27.84 & 26.32 & 43.22 & 50.31 & 31.76
& 18.94 & 17.24 & 43.40 & \textbf{19.84} & 12.58 \\

\midrule
\textbf{HDC-CL (ours)}
& \textbf{31.13} & \textbf{27.34} & \textbf{44.58} & \textbf{54.68} & \textbf{33.09}
& \textbf{19.26} & \textbf{18.45} & \textbf{44.32} & 19.16 & \textbf{13.00} \\
\bottomrule
\end{tabular}
\label{tab55}
\end{table*}


\subsection{Ablation Study}

\textbf{Effects of context distillation losses.}
Table~\ref{table33} presents the joint ablation results of different loss components under the presence of the alignment loss $\mathcal{L}_{\text{align}}$ on both the UAV-SCC\textsubscript{Rich} and UAV-SCC\textsubscript{Simple} datasets. Overall, the results show that the three loss terms provide complementary benefits, and their joint optimization achieves the best overall performance across the two benchmarks.

On UAV-SCC\textsubscript{Rich}, individual loss terms or partial combinations may yield the best results on specific metrics, such as ROUGE-L or BLEU-4. However, the full configuration consistently provides the strongest overall balance, achieving the best METEOR, CIDEr, and SPICE scores. On UAV-SCC\textsubscript{Simple}, the full loss configuration achieves the best performance on all five evaluation metrics, further demonstrating the effectiveness of the proposed multi-level loss design.
These findings indicate that $\mathcal{L}_{\text{global}}$, $\mathcal{L}_{\text{region}}$, and $\mathcal{L}_{\text{HSIC}}$ capture complementary aspects of scene-change learning. Their combination leads to more robust semantic alignment and improves the quality of generated change descriptions.
\textbf{Effects of HCM-OCC.}
Table~\ref{table4} presents the ablation results of the proposed HCM-OCC module on the UAV-SCC\textsubscript{Simple} and UAV-SCC\textsubscript{Rich} datasets. 
Overall, introducing HCM-OCC consistently improves performance across most evaluation metrics, demonstrating its effectiveness in enhancing cross-modal semantic alignment for UAV scene change captioning.

Within the HDC-CL framework, incorporating HCM-OCC significantly improves the CIDEr scores to 54.68 and 19.16 on UAV-SCC\textsubscript{Simple} and UAV-SCC\textsubscript{Rich}, respectively, yielding gains of 3.21 and 0.50 compared with the corresponding variants without HCM-OCC. 
These improvements indicate that the proposed hierarchical orientation consistency calibration effectively strengthens the interaction between visual change features and directional textual semantics, leading to more accurate change descriptions.
Furthermore, applying HCM-OCC to external baselines also leads to consistent improvements. 
For example, integrating HCM-OCC into CARD increases the CIDEr score by 2.80 on UAV-SCC\textsubscript{Rich}, while DIRL also benefits from similar performance gains. 
These results demonstrate that the proposed module is architecture-agnostic and can be readily incorporated into different change captioning frameworks to improve their cross-modal reasoning capability.

\textbf{Effects of DALT.}
To evaluate the effectiveness of the proposed mask generation strategy in DALT, we conduct ablation experiments on both the UAV-SCC\textsubscript{Simple} and UAV-SCC\textsubscript{Rich} datasets, as reported in Table~\ref{tab55}. 
Four variants are compared with the full HDC-CL model: 
(1) \textit{w/o Mask}, which removes region decomposition; 
(2) \textit{Random Mask}, which replaces the semantic mask with randomly generated binary maps; 
(3) \textit{Random Direction Window}, which samples shift directions randomly; and 
(4) \textit{No Direction Window}, which removes the directional window constraint and relies only on global matching.

From the results, all simplified variants lead to performance degradation compared with the full HDC-CL model on both datasets, demonstrating the importance of the proposed mask generation strategy. 
In particular, removing the mask generation mechanism (\textit{w/o Mask}) significantly reduces caption quality, indicating that explicit region decomposition is crucial for capturing meaningful scene variations. 
Using randomly generated masks or randomly sampled shift directions also degrades performance, suggesting that semantically guided region partition and directional constraints are essential for reliable feature correspondence. 
Overall, the full HDC-CL model consistently achieves the best performance across most evaluation metrics on both datasets, confirming that the proposed shift voting mechanism and direction-aware mask generation effectively improve feature alignment and enhance UAV scene change understanding.


\begin{figure}
	\centering
	\includegraphics[width=\columnwidth]{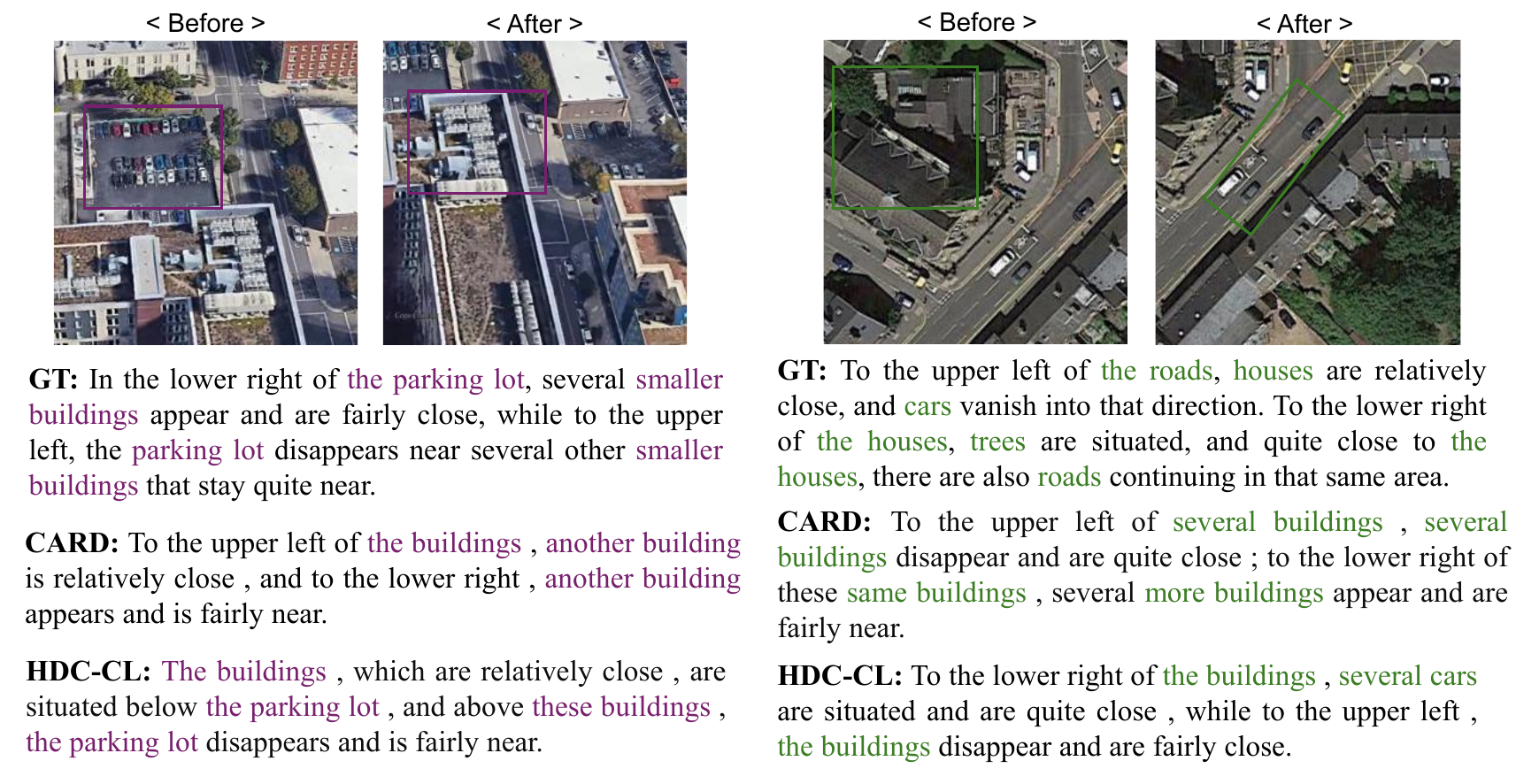}
	\caption{Qualitative comparison of generated captions on the UAV-SCC\textsubscript{Simple} dataset. Text in different colors highlights the described objects or regions.}
	\label{fig:figure4}
\end{figure}

\begin{figure}
  \centering
  \includegraphics[width=\columnwidth]{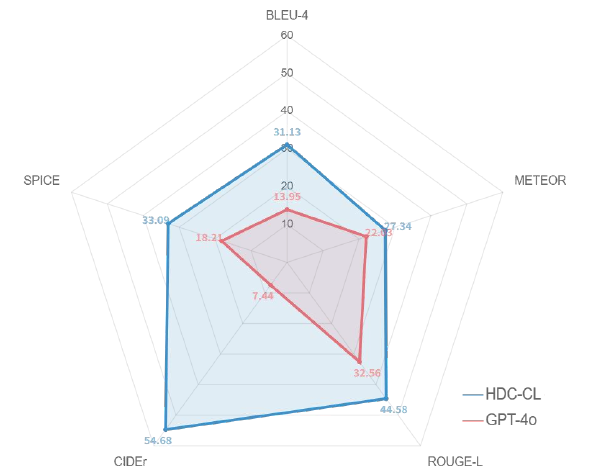}
  \caption{Performance comparison between GPT‑4o and our HDC-CL model on the UAV-SCC\textsubscript{Simple} dataset.}
  \label{mml}
\end{figure}

\textbf{Qualitative Analysis.}
To better understand the effectiveness of the proposed HDC-CL, we compare the scene-change caption visualizations generated by the state-of-the-art method CARD~\cite{tu_contextaware_2024} and the proposed HDC-CL model, as well as the ground truth captions (GT).
As shown in Fig.~\ref{fig:figure4}, we can see the generated captions by HDC-CL are more accurate and more discriminative compared to the state-of-the-art CARD~\cite{tu_contextaware_2024}, which are also consistent with the ground truths. 

Furthermore, driven by exploratory curiosity, we also evaluated the performance of the GPT‑4o model on our dataset. As shown in Figure~\ref{mml}, the results reveal that GPT‑4o performs significantly worse compared to our lightweight task-specific model. This further reinforces the observation that general-purpose large multimodal models (LMMs), although highly capable in open-domain settings, tend to underperform in domain-specific UAV scenarios where the visual and linguistic patterns differ substantially from web-scale training data. Moreover, due to their large memory footprint and high inference latency, such models remain impractical for real-time deployment on UAV platforms or edge-cloud infrastructures with constrained resources.

\section{Conclusion}
In this paper, we introduced UAV scene change captioning, a novel task to describe semantic changes in dynamic aerial imagery captured from a movable viewpoint.
To tackle the challenges caused by temporal–spatial scene variation, we propose the HDC-CL method, which integrates a Dynamic Adaptive Layout Transformer (DALT) and a Hierarchical Cross-modal Orientation Consistency Calibration (HCM-OCC) strategy to adapt to dynamic spatial layouts and enhance cross-modal semantic alignment, respectively. In particular, DALT is equipped with a shift voting mechanism to achieve robust feature-level alignment under viewpoint-induced parallax.
To further evaluate HDC-CL, we construct a benchmark dataset, named UAV-SCC, which includes two annotation variants tailored for different levels of scene understanding and descriptive complexity. We conduct extensive experiments to demonstrate the effectiveness of HDC-CL, achieving substantial and consistent performance gains over existing approaches. We believe this dataset, together with the newly defined UAV scene change captioning task, will further facilitate future research.




\clearpage 





\appendix
\section{Dataset Details}
\label{appendixA}
\subsection{Dataset Construction Process}

To construct the \textbf{UAV Scene Change Captioning(UAV-SCC) Dataset}, we focus on two key aspects: image pair generation and caption annotation. The overall data construction process is described as follows.

For the creation of image pairs, we utilized UAV imagery from two publicly available datasets, \textit{GeoText-1652} and \textit{UAVDT}. Since these datasets were not originally designed for scene-change tasks, we first extracted raw UAV images and reorganized them into scene-change pairs by identifying frames that share partially common regions while exhibiting noticeable variations such as viewpoint shifts, object appearance or disappearance, and illumination or scale differences. To improve scene diversity, images from both sources were jointly considered, enabling complementary coverage of urban, suburban, and traffic-related environments. All selected image pairs were further processed using geometric alignment procedures to enhance overall spatial consistency across the dataset.

Regarding caption annotation, three domain experts annotated scene change descriptions for consecutive pre-change and post-change images.  The annotators were required to employ diverse linguistic expressions when describing identical scene changes.  To ensure annotation consistency, all experts underwent unified training prior to the task and were provided with comprehensive annotation guidelines accompanied by illustrative examples. In addition, for each image pair, captions were also annotated in the reverse temporal direction (i.e., from after to before), resulting in bidirectional textual descriptions.  All subsequent model evaluations were conducted using an equal amount of data to ensure fair comparison and consistency.

The UAV-SCC dataset is released in two versions, which differ in scene complexity, linguistic characteristics, and annotation density.

\begin{itemize}
    \item \textbf{UAV-SCC\textsubscript{Simple}}: includes \textbf{9,017} image pairs with relatively clear and easily recognizable scene differences. Each image pair is annotated with \textbf{three reference captions}, which generally adopt simpler sentence structures and more direct expressions, focusing mainly on describing observable scene changes without detailed mentions of object categories or attributes.

    \item \textbf{UAV-SCC\textsubscript{Rich}}: contains \textbf{7,054} image pairs featuring more complex spatial layouts and subtle visual variations. Each image pair is annotated with \textbf{five reference captions}, providing denser and more diverse linguistic descriptions. The captions in this version are linguistically richer and semantically finer-grained, often involving a wider variety of objects, their spatial relationships, and descriptive attributes.
\end{itemize}

These two versions jointly provide a comprehensive benchmark for evaluating scene change captioning models under different levels of visual complexity and linguistic granularity.

\begin{figure*}
  \centering
  \includegraphics[width=\textwidth]{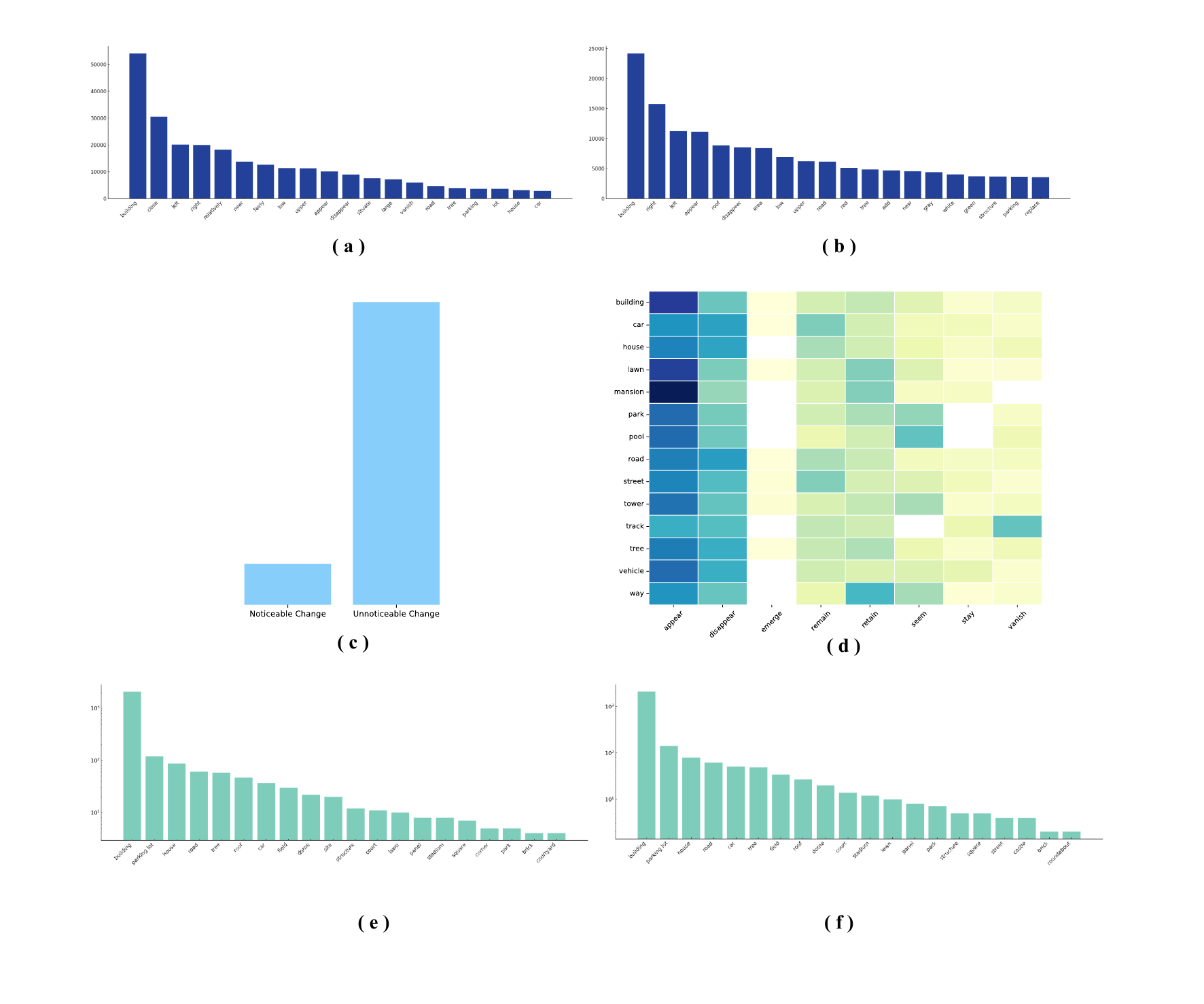}
  \caption{Statistical and linguistic analyses of the UAV-SCC dataset.
(a) Word frequency distribution of captions in the UAV-SCC\textsubscript{Simple} dataset.
(b) Word frequency distribution of captions in the UAV-SCC\textsubscript{Rich} dataset.
(c) Distribution of noticeable and unnoticeable scene changes.
(d) Heatmap of noun–verb co-occurrence frequencies, illustrating how object categories associate with specific change verbs.
(e) Object-noun frequency distribution in the test-set captions generated by HDC-CL.
(f) Object-noun frequency distribution in the test-set captions generated by CARD.}
  \label{spp1}
\end{figure*}


\begin{table*}[t]
\centering
\footnotesize
\caption{
Object-level prediction statistics for CARD and HDC-CL.
``Pred-A'' / ``Pred-D'' denote the number of images where the model predicts
object appearance or disappearance; ``Pred-Desc'' denotes cases where the
object is mentioned without a change verb. ``Corr-A'' / ``Corr-D'' count cases
where appearance or disappearance predictions match the ground truth.
``Co-Mentioned'' counts images where both prediction and ground truth mention
the object (regardless of change status). ``Accuracy'' is computed as:
(Corr-A + Corr-D + Co-Mentioned) / (Pred-A + Pred-D + Pred-Desc).
}
\begin{tabular}{l|c|ccc|cc|c|c}
\toprule
Object & Model &
Pred-A & Pred-D & Pred-Desc &
Corr-A & Corr-D &
Co-Mentioned & Accuracy (\%) \\
\midrule
\multirow{2}{*}{Car}
& CARD   & 1 & 5 & 24 & 1 & 0 & 8 & 30.0 \\
& HDC-CL & 6 & 4 & 14 & 2 & 0 & 6 & 33.3 \\
\midrule
\multirow{2}{*}{Parking lot}
& CARD   & 0 & 5 & 56 & 0 & 0 & 25 & 41.0 \\
& HDC-CL & 0 & 6 & 46 & 0 & 2 & 22 & 46.2 \\
\midrule
\multirow{2}{*}{Road}
& CARD   & 19 & 13 & 3 & 1 & 1 & 3 & 14.3 \\
& HDC-CL & 23 & 12 & 17 & 4 & 1 & 8 & 25.0 \\
\midrule
\multirow{2}{*}{Tree}
& CARD   &  1 & 14 & 9 & 0 & 3 & 4 & 29.2 \\
& HDC-CL & 1 & 19 & 3 & 1 & 0 & 5 & 26.1 \\
\bottomrule
\end{tabular}
\label{table6}
\end{table*}

\begin{figure*}[t]
  \centering
  \includegraphics[width=\textwidth]{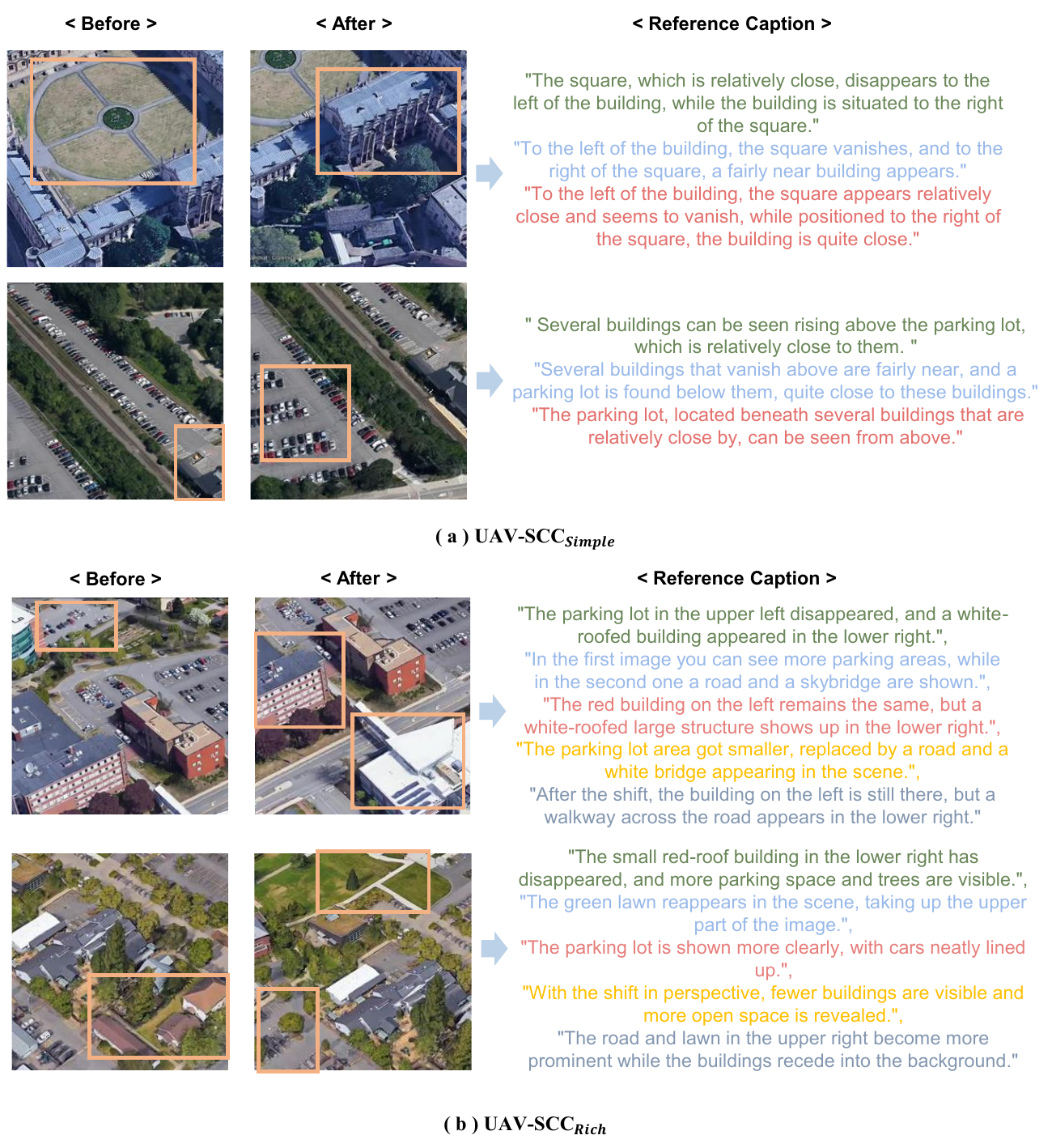}
  \caption{Examples of image pairs and corresponding reference captions in the UAV-SCC dataset. (a) \textbf{UAV-SCC\textsubscript{Simple}}: each image pair shows clear and easily distinguishable scene changes, and the captions mainly provide straightforward descriptions with simple sentence structures. (b) \textbf{UAV-SCC\textsubscript{Rich}}: the scenes exhibit more complex spatial layouts and finer-grained variations, while the captions include richer linguistic details, describing diverse objects, attributes, and spatial relationships.}
  \label{spp3}
\end{figure*}

\subsection{Dataset Analyses of UAV-SCC Dataset Benchmark}

We further conduct comprehensive statistical and visualization analyses on our constructed UAV-SCC dataset to demonstrate its linguistic characteristics and scene diversity.
Specifically, we perform a detailed statistical analysis of the generated captions across the two dataset versions.
After removing common stop words, we calculate the word frequency distributions of scene-related nouns and change-related verbs to better understand the descriptive focus and linguistic patterns.
As shown in Figure~\ref{spp1}(a), the UAV-SCC\textsubscript{Simple} dataset is dominated by the noun \emph{building}, whose frequency is significantly higher than all other tokens. Besides this, the top-ranked words are mainly proximity- and layout-related terms (e.g., “close”, “right”, “near”, “relatively”, “fairly”, “few”), together with a smaller set of object nouns such as \emph{cars}, \emph{trees}, \emph{road}, \emph{parking lot}, and \emph{houses}. This distribution indicates that captions in the Simple type tend to focus on a few salient scene elements and their coarse spatial relations, rather than using a wide variety of object categories.
In contrast, Figure~\ref{spp1}(b) presents the word frequency distribution for the UAV-SCC\textsubscript{Rich} dataset. The vocabulary in this version is more diverse, involving a wider range of scene components and object attributes. High-frequency words such as “building,” “road,” “roof,” “tree,” “gray,” and “green” suggest that captions not only describe object existence and spatial relations but also incorporate color and appearance attributes, reflecting more fine-grained and descriptive language.
This richer lexical diversity highlights that the UAV-SCC\textsubscript{Rich} subset captures more complex scene semantics and varied linguistic expressions, which are crucial for evaluating a model’s capability to generate detailed and semantically grounded change descriptions.
Additionally, we count the number of noticeable changes and unnoticeable changes across the dataset (Figure~\ref{spp1}(c)). The results show that while the dataset contains a certain proportion of significant scene changes, it also includes a large number of subtle changes, increasing the difficulty and diversity of the scene-change captioning task.
Finally, we analyze the co-occurrence relationships between nouns and verbs and visualize them as a heatmap of noun-verb pairs (Figure~\ref{spp1}(d)). The results demonstrate that different object categories (e.g., building, car, house, road, etc.) exhibit clear collocation patterns with specific change verbs. For example, "building" is often associated with "appear" or "disappear", while "road" tends to co-occur with "remain" or "retain", reflecting the regularity of change expressions in our dataset and the realism of scene content.

We observed an interesting phenomenon: 
(a) similar to other real-world scene datasets, Figure~\ref{spp1}(a) exhibits a long-tail distribution. This is mainly due to certain frequently occurring objects in real scenes—particularly large objects such as building—that tend to dominate the visual field. 
(b) We further analyze the object-noun distributions in the model-generated captions, as shown in Figures~\ref{spp1}(e) and \ref{spp1}(f). To better reveal distributional differences, both plots use logarithmic scaling on the vertical axis rather than raw counts, which allows low-frequency categories to remain visible despite the inherently imbalanced nature of real-world scenes.
Figure~\ref{spp1}(e) shows that captions generated by HDC-CL exhibit a smoother long-tail distribution. While high-frequency objects such as building, parking lot, and road naturally dominate due to their prevalence in aerial imagery, HDC-CL still produces a non-negligible number of mentions for low-frequency categories.
This indicates that our method maintains better sensitivity to diverse scene elements, even those that appear infrequently in the training data.
In contrast, Figure~\ref{spp1}(f) reveals that CARD displays a more skewed distribution, heavily biased toward the most frequent nouns while rarely mentioning low-frequency ones. This suggests a stronger tendency to overfit high-frequency categories and weaker generalization toward less prominent or small-scale objects.
A possible reason is that our method shows higher sensitivity to diverse scenes and objects, including those with low occurrence frequency.
As illustrated in Figure~\ref{spp_6} (heatmap visualization), this sensitivity helps mitigate the impact of the long-tail effect to some extent.

Table~\ref{table6} presents an object-level comparison between CARD and HDC-CL based on their alignment with the ground-truth captions. It is worth noting that a generated description may still be semantically correct even if it does not exactly match the annotated phrasing; thus, this analysis reflects agreement with the annotation style rather than absolute factual correctness.
Under this metric, HDC-CL generally achieves stronger alignment with the ground-truth patterns. For objects such as car and parking lot, HDC-CL produces more correct appearance/disappearance predictions (Corr-A / Corr-D) and generates fewer unmatched mentions, resulting in higher overall consistency—33.3\% vs. 30.0\% for car, and 46.2\% vs. 41.0\% for parking lot. These gains indicate that HDC-CL more closely adheres to the descriptive tendencies encoded in UAV-SCC.
Although the evaluation does not account for all factually plausible descriptions, the improvements observed for HDC-CL suggest that its generated captions better reflect both the visual content and the annotation conventions of the dataset.

In addition to the quantitative analysis of word frequencies and linguistic structures, we further provide several representative visualization examples from both UAV-SCC\textsubscript{Simple} and UAV-SCC\textsubscript{Rich} to illustrate the semantic alignment between image pairs and their corresponding captions, as shown in Figure~\ref{spp3}.
For each dataset, we present “before” and “after” images along with multiple reference captions that describe the observed scene changes.
Figure~\ref{spp3}(a) shows examples from UAV-SCC\textsubscript{Simple}, where the scene differences are relatively clear and the captions provide concise descriptions of noticeable changes such as the appearance or disappearance of buildings and the reconfiguration of parking areas.
In contrast, Figure~\ref{spp3}(b) presents examples from UAV-SCC\textsubscript{Rich}, which involve more complex spatial layouts and finer-grained variations. The captions in this version describe a broader range of object categories, spatial relations, and appearance attributes, reflecting higher linguistic diversity and semantic detail.
These visualization examples demonstrate that the UAV-SCC dataset effectively captures diverse change scenarios and provides high-quality textual annotations that facilitate the study of scene-change understanding and language generation.

\section{Shift Voting Mechanism}
\label{sec:appendix_shiftvoting}

This section provides additional implementation details of the 
\textit{shift voting mechanism} described in the main paper.
Given the flattened token sequences 
$\mathbf{I}_{\text{bef}}=\{\mathbf{I}_{\text{bef}}[i]\}_{i=1}^{L}$ 
and 
$\mathbf{I}_{\text{aft}}=\{\mathbf{I}_{\text{aft}}[j]\}_{j=1}^{L}$,
we first $\ell_2$-normalize each token and compute the pairwise similarities as
\begin{equation}
    s_{ij}=\frac{\mathbf{I}_{\text{bef}}[i]^\top\mathbf{I}_{\text{aft}}[j]}{\tau},
\label{eq:sim_appendix}
\end{equation}
where the temperature $\tau$ is initialized to $0.07$.
For each $\mathbf{I}_{\text{bef}}[i]$, we retain only its most similar counterpart 
in $\mathbf{I}_{\text{aft}}$ for stable displacement estimation, and compute the
discrete shift $\Delta_{ij}=\mathbf{p}_j-\mathbf{p}_i$ based on their spatial coordinates.

Following the formulation in the main paper, we build a weighted displacement-voting map
\begin{equation}
    S(\Delta)=\sum_{i,j}\max(s_{ij},0)\,\mathbf{1}(\Delta_{ij}=\Delta),
\label{eq:voting_appendix}
\end{equation}
and determine the dominant geometric offset by
\begin{equation}
    \Delta^\star=\arg\max_{\Delta} S(\Delta).
\label{eq:dominant_appendix}
\end{equation}

To further ensure spatial coherence, we keep only correspondences
satisfying $\|\Delta_{ij}-\Delta^\star\|_\infty\le R$, removing geometrically inconsistent matches. In practice, we empirically set $R=2$ for all experiments.
A Mutual Nearest Neighbor (MNN) criterion is also enforced, keeping a pair $(i,j)$ only if they are 
mutual nearest neighbors and their similarity exceeds a predefined threshold.
After directional filtering and MNN validation, we obtain a reliable binary indicator for each token,
which is finally reshaped to the spatial layout to form the 2D overlap mask used in the main framework.

\begin{table}
    \centering
    \fontsize{9}{11}\selectfont
    \renewcommand\tabcolsep{5pt}
    \caption{Comparison of image vs. text processing in terms of latency and data size (based on Matrice 4D specifications).}
    \begin{tabular}{l|l|l}
        \toprule
        \textbf{Period} & \textbf{Estimated Latency} & \textbf{Data Size} \\
        \midrule
        Image Transmission & $\approx$ 0.6 -- 2 seconds & $\approx$ 10 MB \\
        Text Transmission  & $\approx$ 5 -- 10 ms       & $<$ 1 KB \\
        Text Inference & $\approx$ 77 ms           & --- \\
        Text Inference + Transmission & $\approx$ 82 -- 87 ms & $<$ 1 KB \\
        \bottomrule
    \end{tabular}
    \label{table_latency_comparison}
\end{table}

\begin{table*}[t]
    \centering
    \fontsize{9}{11}\selectfont
    \caption{Effects of $\lambda$ on UAV-SCC\textsubscript{Simple} and UAV-SCC\textsubscript{Rich}.}
   
    \setlength{\tabcolsep}{3pt}
    \begin{tabular*}{\textwidth}{@{\extracolsep{\fill}} c|ccccc|ccccc}
        \toprule
        \multirow{2}{*}{$\lambda$} 
        & \multicolumn{5}{c|}{\textbf{UAV-SCC\textsubscript{Simple}}} 
        & \multicolumn{5}{c}{\textbf{UAV-SCC\textsubscript{Rich}}} \\
        & B & M & R & C & S 
        & B & M & R & C & S \\
        \midrule
        0    & 29.26 & 27.18 & 44.33 & 49.55 & 32.56 
             & \textbf{19.46} & 16.51 & \textbf{45.29} & 13.86 & 12.60 \\

        0.01 & \textbf{31.13} & 27.34 & \textbf{44.58} &\textbf{ 54.68} & 33.09
             & 19.26 & \textbf{18.45} & 44.32& \textbf{19.16} & 13.00 \\

        0.02 &29.01 & 26.69 & 43.02 & 46.80 & 31.45
             & 18.92 & 17.65 & 43.85 & 18.27 & 13.02 \\

        0.03 & 29.97 & 27.28 & 44.16 & 53.81 & \textbf{33.13}
             & 18.88 & 17.99 & 44.59 & 18.78 & \textbf{13.09} \\

        0.04 & 29.02 & 27.35 & 43.86 & 48.45 & 32.89
             & 18.09 & 16.81 & 43.88 & 18.09 & 12.20 \\

        0.05 & 29.67 & \textbf{27.44} & 43.96 & 51.33 & 32.36
             & 18.79 & 17.43 & 44.45 & 19.31 & 12.84 \\
        \bottomrule  
    \end{tabular*}
    \label{table_ssp2}
\end{table*}

\section{Caption Generation}
In order to better explain the process of caption generation, here is a further explanation of how to convert the visual difference representation into natural language patterns of scene changes.

Starting with the unified difference representation 
$\mathbf{D} \in \mathbb{R}^{N \times d}$, 
we use a Transformer decoder to sequentially generate words that describe the directional change. 
For each ground-truth caption of length $m$, 
we obtain its token embeddings 
$\mathbf{E} \in \mathbb{R}^{m \times d}$, 
and add positional encodings to preserve the order of words in the sequence.

These embeddings are first processed by a masked self-attention module to capture dependencies within the sentence. This operation ensures that, during training, each word attends only to preceding tokens, facilitating autoregressive decoding. The output of this step can be formulated as:
\begin{equation}
\hat{\mathbf{E}} = \mathrm{LayerNorm}\left( \mathbf{E} + \mathrm{SelfAttn}(\mathbf{E}) \right),
\label{eq:selfattn_layernorm}
\end{equation}
where the self-attention is applied in a masked fashion, and layer normalization is used to stabilize training.

Next, the decoder integrates visual information by attending to the difference representation \textbf{D} through a multi-head cross-attention mechanism. This allows each word position to align with the most relevant visual changes:
\begin{equation}
\tilde{\mathbf{H}} = \mathrm{LayerNorm}\left( \hat{\mathbf{E}} + \mathrm{CrossAttn}\left( \hat{\mathbf{E}}, \mathbf{D} \right) \right) ,
\label{eq:crossattn_layernorm}
\end{equation}

Following this, a position-wise feed-forward network is applied to further transform the attended features, enhancing their expressive power:
\begin{equation}
\mathbf{\hat{H}} = \mathrm{LayerNorm}\left( \tilde{\mathbf{H}} + \mathrm{FFN}\left( \tilde{\mathbf{H}} \right) \right),
\label{eq:ffn_layernorm}
\end{equation}

The final output 
$\mathbf{\hat{H}} \in \mathbb{R}^{m \times d}$ 
represents the contextualized token states, incorporating both linguistic and visual cues. 
Each of these states is projected to a vocabulary distribution using a linear layer followed by a softmax:
\begin{equation}
\mathbf{S} = \mathrm{Softmax}\left( \mathbf{\hat{H}} \mathbf{W}_s + \mathbf{b}_s \right).
\label{eq:softmax_projection}
\end{equation}
where $\mathbf{W}_s \in \mathbb{R}^{d \times u}$ and $\mathbf{b}_s \in \mathbb{R}^u$ are learnable parameters, with $u$ being the vocabulary size.

Through this decoding process, the model generates captions that are not only fluent and coherent but also closely aligned with the underlying scene changes captured in the visual features.

\section{Experiment}
\label{appendixB}
\subsection{Implementation Details}
For fair comparison, we follow previous multi-change captioning methods and adopt a pre-trained ResNet-101 to extract local features from the input image pairs on the UAV-SCC dataset. 
The temperature $\tau$ used in Eqs.~(5) and (6) is set to 0.07. 
The margin hyperparameter $\gamma$ in Eq.~(13) is empirically set to 0.3.
The model is implemented in PyTorch and trained on a single RTX 3090 GPU using the Adam optimizer to minimize the negative log-likelihood loss defined in Eq.~(15) of the main paper.
All experiments are conducted on a single NVIDIA RTX 3090 GPU under a single-task training setting.
The total training time for one full run on the UAV-SCC\textsubscript{Simple} dataset is approximately 61 minutes, with a peak GPU memory usage of 4.95 GB,
while training on the UAV-SCC\textsubscript{Rich} dataset takes about 55 minutes and uses up to 4.50 GB of GPU memory.


\begin{figure*}
  \centering
  \includegraphics[width=\textwidth]{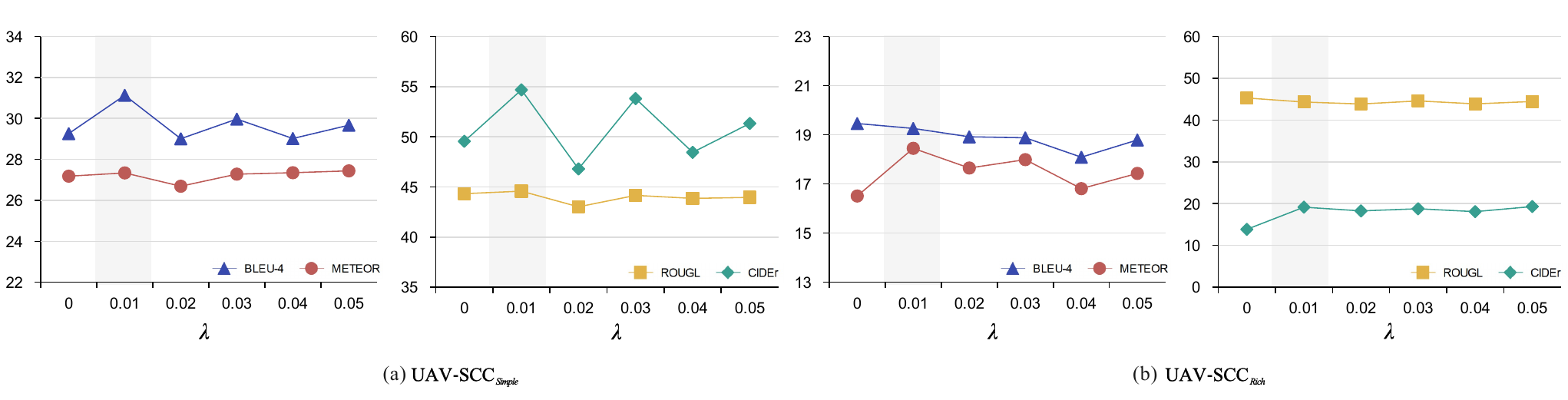}
  \caption{Quantitative analysis of the impact of the hyperparameter $\lambda$ on model performance.
(a) UAV-SCC\textsubscript{Simple} and (b) UAV-SCC\textsubscript{Rich} show the variation of different evaluation metrics (BLEU4, METEOR, ROUGE-L, and CIDEr) with respect to $\lambda$.
A moderate value of $\lambda$ (around 0.01) generally yields better overall performance, while excessively large values lead to minor fluctuations or degradation, indicating the need for a balanced weighting among loss components.}
  \label{spp4}
\end{figure*}

\subsection{Model Analysis}
\paragraph{Latency analysis}
As shown in Table~\ref{table_latency_comparison}, the total time required for transmitting images (approximately 10 MB) ranges from 0.6 to 2 seconds, depending on the available bandwidth and transfer conditions. In contrast, the transmission of a short text message (less than 1 KB) via the low-latency control link only takes about 5–10 milliseconds. Combined with the model’s inference time of around 77 milliseconds, the total latency for text generation and transmission is approximately 82-87 milliseconds.
This comparison clearly demonstrates that our text-based approach achieves significantly lower overall latency compared to transmitting raw image data. The reduced data size and the efficient use of control link bandwidth make this method particularly well-suited for applications on UAV platforms, especially in scenarios where communication bandwidth is limited or fast decision-making is required.


\begin{figure*}
  \centering
  \includegraphics[width=\textwidth]{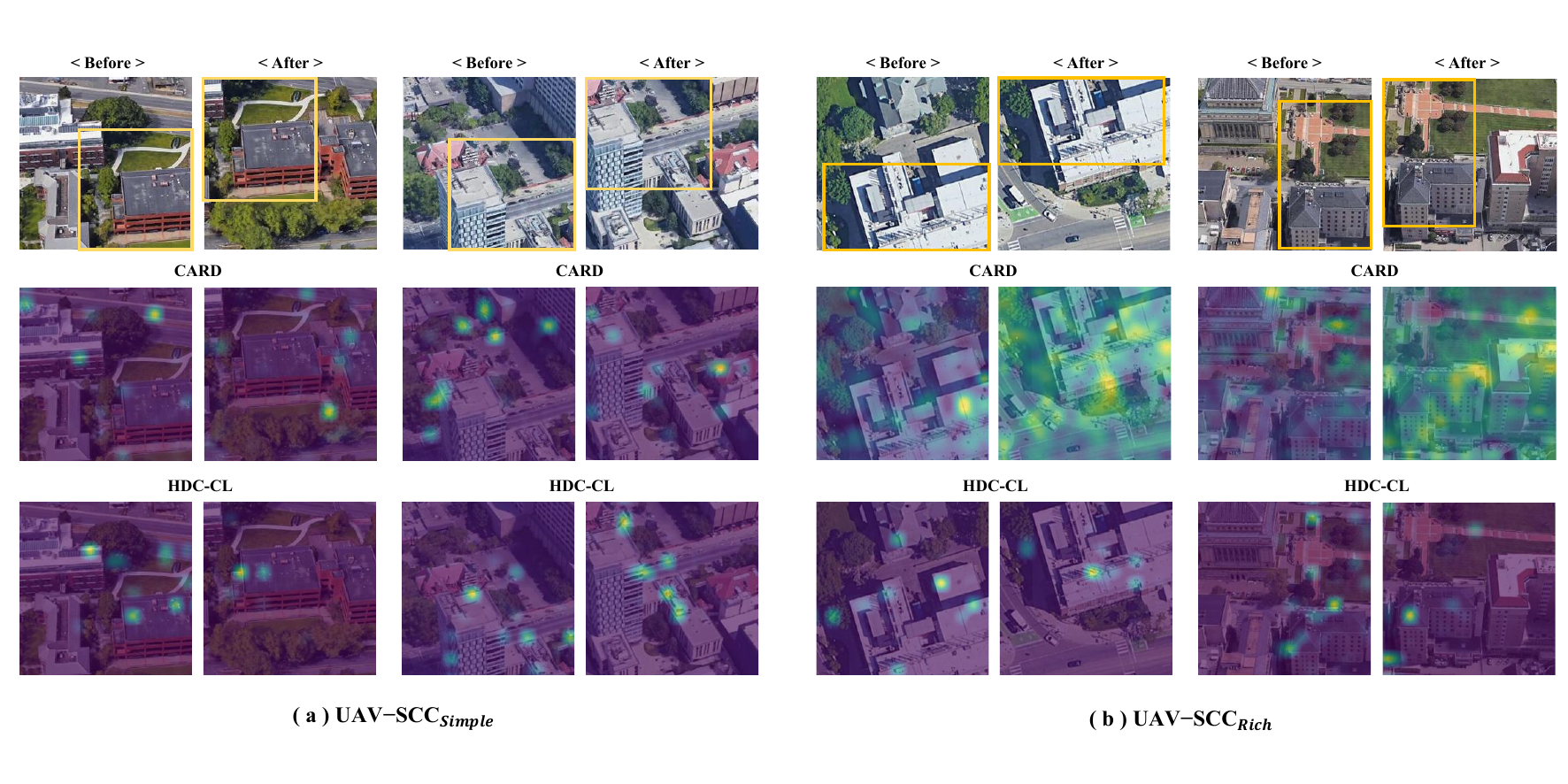}
  \caption{Qualitative comparison of attention maps from CARD and our HDC-CL. The yellow box marks the approximate common region between the image pair; it is manually added for visualization and is not ground-truth annotation. HDC-CL yields more accurate and focused change responses than CARD.}
  \label{spp_6}
\end{figure*}

\begin{figure*}
  \centering
  \includegraphics[width=\textwidth]{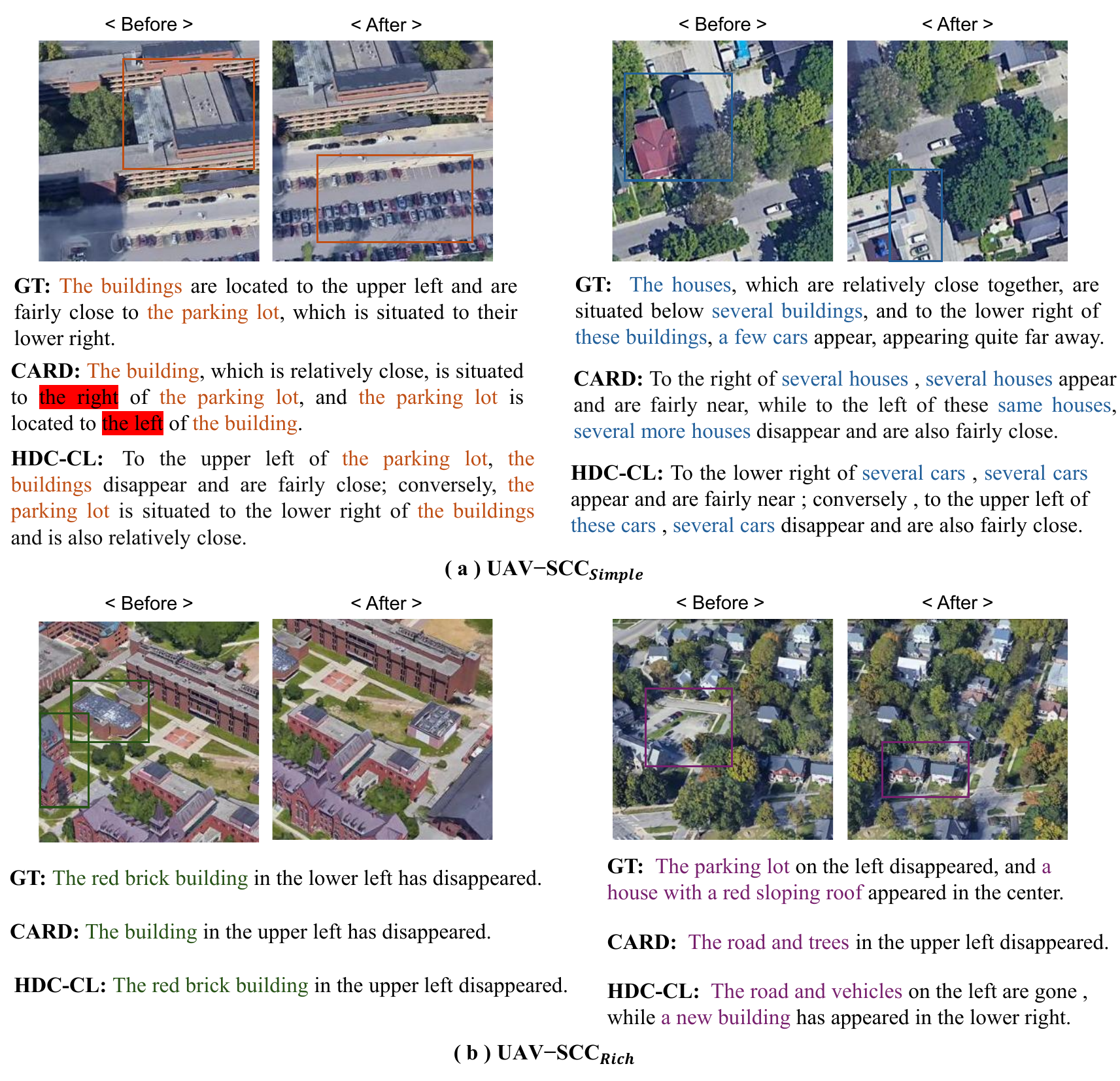}
  \caption{Qualitative captioning examples on (a)UAV-SCC\textsubscript{Simple} and (b)UAV-SCC\textsubscript{Rich}. For each image pair, we show the ground-truth caption (GT), the output from CARD, and the caption generated by our proposed HDC-CL. HDC-CL produces change descriptions that are more consistent with the visual evidence and spatial relations in the before/after images, while CARD often yields incorrect object references or misaligned spatial descriptions.
}
  \label{spp8}
\end{figure*}

\paragraph{Impact of the hyperparameters.}
We further investigate the effect of the hyperparameter $\lambda$ in our HDC-CL model on both UAV-SCC\textsubscript{Simple} and UAV-SCC\textsubscript{Rich} datasets.
Table~\ref{table_ssp2} summarizes the quantitative results under different $\lambda$ values.
This parameter controls the relative weighting among multiple loss components in the overall optimization objective, thereby influencing the balance between the main caption generation loss and auxiliary regularization terms.
As shown in the results, introducing a moderate weighting factor ($\lambda>0$) generally leads to improved performance compared to removing it ($\lambda=0$), suggesting that incorporating auxiliary constraints can enhance representation learning and promote better semantic alignment.
However, the improvement is not uniform across all metrics or datasets, as overly large $\lambda$ values may cause certain losses to dominate the optimization process, leading to fluctuations or slight degradation in captioning quality.
To provide a clearer view of these trends, Figure~\ref{spp4} visualizes the variation of different evaluation metrics with respect to $\lambda$.
The figure confirms that moderate values (around $\lambda=0.01$) yield the optimal balance between constraint regularization and descriptive accuracy.
Therefore, we adopt $\lambda=0.01$ as the default setting in all subsequent experiments.

\begin{table*}[t]
    \centering
    \fontsize{9}{11}\selectfont
    \renewcommand\tabcolsep{6pt}
    \caption{
Joint ablation of context distillation losses on UAV-SCC\textsubscript{Rich} and UAV-SCC\textsubscript{Simple} 
\textbf{without} the alignment loss $\mathcal{L}_{\text{align}}$.
\textbf{Bold} numbers indicate the best performance in each column.
}
    \label{table_loss_joint_noalign}
    
    \begin{tabular}{c c c | c c c c c | c c c c c}
        \toprule
        \multirow{2}{*}{$\mathcal{L}_{\text{global}}$} &
        \multirow{2}{*}{$\mathcal{L}_{\text{region}}$} &
        \multirow{2}{*}{$\mathcal{L}_{\text{HSIC}}$} &
        \multicolumn{5}{c|}{\textbf{UAV-SCC\textsubscript{Rich}}} &
        \multicolumn{5}{c}{\textbf{UAV-SCC\textsubscript{Simple}}}
        \\
        \cmidrule(lr){4-8} \cmidrule(lr){9-13}
        & & &
        \textbf{B} & \textbf{M} & \textbf{R} & \textbf{C} & \textbf{S} &
        \textbf{B} & \textbf{M} & \textbf{R} & \textbf{C} & \textbf{S}
        \\
        \midrule
        $\times$ & $\times$ & $\times$ &
        19.46 & 16.51 & 45.29 & 13.86 & 12.60 &
        29.26 & \textbf{27.18} & 44.33 & 49.55 & \textbf{32.56}
        \\
        $\checkmark$ & $\times$ & $\times$ &
        18.48 & 17.29 & 43.32 & 18.25 & 12.74 &
        28.59 & 26.69 & 42.89 & \textbf{51.67} & 32.16
        \\
        $\times$ & $\checkmark$ & $\times$ &
        18.42 & 17.36 & 43.71 & 18.49 & 12.59 &
        \textbf{29.92} & 26.86 & 44.19 & 50.55 & 31.85
        \\
        $\times$ & $\times$ & $\checkmark$ &
        19.62 & 16.71 & \textbf{45.81} & 16.74 & 12.64 &
        29.87 & 26.66 & \textbf{44.52} & 49.00 & 31.69
        \\
        $\checkmark$ & $\checkmark$ & $\times$ &
        19.67 & 17.68 & 44.51 & 17.63 & \textbf{12.80} &
        28.76 & 26.75 & 43.78 & 51.24 & 32.10
        \\
        $\checkmark$ & $\times$ & $\checkmark$ &
        19.30 & 17.55 & 44.26 & 18.46 & 12.18 &
        28.62 & 26.90 & 43.43 & 50.87 & 32.31
        \\
        $\times$ & $\checkmark$ & $\checkmark$ &
        \textbf{19.85} & \textbf{17.88} & 43.84 & 17.70 & 12.14 &
        29.48 & 26.92& 43.94 & 50.89 & 31.57
        \\

        $\checkmark$ & $\checkmark$ & $\checkmark$ &
        18.86 & 16.99 & 44.03 & \textbf{18.66} & 12.42 &
       29.62 & 26.84 &43.33&
        51.47 & 31.82
        \\

        \bottomrule
    \end{tabular}
\end{table*}

\textbf{Effects of context distillation losses without the alignment loss.}
Table~\ref{table_loss_joint_noalign} reports the joint ablation results of the context distillation losses on both UAV-SCC\textsubscript{Rich} and UAV-SCC\textsubscript{Simple} when the alignment loss $\mathcal{L}_{\text{align}}$ is disabled. Several consistent patterns can be observed across the two datasets.

First, using any single loss component generally leads to moderate improvements, but the effects vary. On UAV-SCC\textsubscript{Rich}, $\mathcal{L}_{\text{HSIC}}$ alone yields the highest ROUGE-L score (45.81), indicating its strong contribution to enhancing the discriminability between common and difference features. Meanwhile, the region-level loss $\mathcal{L}_{\text{region}}$ provides stable gains in CIDEr on both datasets, highlighting the importance of localized supervision even without explicit alignment constraints. The global caption loss $\mathcal{L}_{\text{global}}$ improves CIDEr on UAV-SCC\textsubscript{Rich} and achieves competitive performance on the Simple split, demonstrating its benefit in stabilizing holistic caption generation.
When two losses are combined, the performance becomes more balanced. In particular, the combination of $\mathcal{L}_{\text{global}}$ and $\mathcal{L}_{\text{region}}$ produces the best SPICE on the Rich dataset and strong CIDEr on the Simple dataset (51.24). This suggests that global–regional consistency plays a key role when explicit alignment cues are absent.

Notably, the best overall performance is not achieved when all three losses are combined. Instead, the variant with $\mathcal{L}_{\text{region}}$ and $\mathcal{L}_{\text{HSIC}}$ (without $\mathcal{L}_{\text{global}}$) achieves the highest BLEU-4 (19.85) and METEOR (17.88) on the Rich dataset, while the three-loss configuration performs competitively but not dominantly. This trend indicates that in the absence of alignment supervision, strong global constraints may interfere with fine-grained relational modeling, whereas the region-level and decorrelation objectives form a more compatible combination.
Overall, these results reveal that the proposed loss components remain effective even without $\mathcal{L}_{\text{align}}$, but their interactions differ from the aligned setting. The findings highlight the complementary roles of region-level grounding and feature decorrelation in supporting robust scene change captioning when explicit alignment cues are unavailable.

\paragraph{Case study.}

Figure~\ref{spp_6} presents qualitative attention visualizations for CARD and our HDC-CL on representative examples from both UAV-SCC\textsubscript{Simple} and UAV-SCC\textsubscript{Rich}. The yellow box in the top row marks the approximate common region shared between each before–after pair; it is manually added solely for visualization and is not a ground-truth annotation. All attention responses are computed using the cross-view interaction defined in Eq.~(9) of the main paper.
Across the examples, CARD frequently produces diffuse or incorrectly distributed activations, often spreading outside the true common regions and failing to maintain stable correspondence under viewpoint changes or scene clutter. In contrast, HDC-CL consistently yields sharper and more localized responses, with high-attention areas concentrated within the actual overlapping regions of the two views. This demonstrates that our hierarchical decompositional calibration effectively suppresses irrelevant differences while accurately capturing cross-view shared content.
Notably, the advantage of HDC-CL persists in the more complex UAV-SCC\textsubscript{Rich} examples, where richer scene structures and fine-grained variations further challenge correspondence estimation. Even under these conditions, HDC-CL maintains stable activation patterns aligned with the true common regions, whereas CARD’s responses become noticeably scattered. These results confirm the robustness of our attention mechanism.

As shown in Figure~\ref{spp8}, CARD produces inaccurate or inconsistent change descriptions, especially in spatial relations. In contrast, HDC-CL tends to generate captions that are more faithful to the actual visual changes in both the UAV-SCC\textsubscript{Simple} and UAV-SCC\textsubscript{Rich} datasets.


\printcredits

\bibliographystyle{cas-model2-names}

\bibliography{cas-refs}



\end{document}